\documentclass{article}

\IfFileExists{VLA-AN-style.sty}{\usepackage{./VLA-AN-style}}{\usepackage[a4paper,margin=24mm]{geometry}}
\usepackage{amsmath,amssymb,booktabs,graphicx,hyperref,url,cite}
\usepackage{placeins,tikz,array,longtable,tabularx,xcolor}
\usepackage{float,array}
\usetikzlibrary{arrows.meta,fit,positioning,calc}
\providecommand{\keywords}[1]{\par\noindent\textbf{Keywords:} #1\par}

\author{}\date{}
\hypersetup{hidelinks,pdftitle={DiffWAM: Accelerating World-Action Models for Continuous UAV Navigation},
pdfauthor={Mo Zhu et al.},pdfkeywords={latent-to-pose,video world models,metric geometry,UAV navigation}}
\makeatletter
\renewcommand{\normalsize}{\@setfontsize\normalsize{11}{14}
\abovedisplayskip 9pt plus 2pt minus 2pt
\belowdisplayskip 9pt plus 2pt minus 2pt
\abovedisplayshortskip 3pt plus 2pt
\belowdisplayshortskip 6pt plus 2pt}
\makeatother
\DeclareMathSizes{11}{11}{8}{6}
\AtBeginDocument{\normalsize\raggedbottom}
\begin{document}
\maketitle
\begin{abstract}

Pretrained video foundation models encode rich semantic and spatiotemporal priors for embodied navigation, yet converting these priors into UAV motion typically requires expensive future-video synthesis and geometric reconstruction. We investigate whether the motion implicit in future visual prediction can instead be recovered directly from the predictive representations of a frozen video model. To this end, we present \textbf{DiffWAM}, a geometry-conditioned navigation world-action model that directly transforms multi-level predictive features into continuous camera trajectories. Its Grid-Motion module preserves spatial-temporal motion associations, while Latent2Pose grounds them with first-frame geometry to recover metrically meaningful 3D motion. Complete video rollouts and geometric reconstruction are required only for offline supervision, eliminating future-video decoding and multi-frame reconstruction during deployment. We further introduce \textbf{FastDreamer}, which overlaps predictive and geometric computation with ongoing flight and performs timestamp-aware asynchronous trajectory handoff for continuous UAV execution.
DiffWAM achieves a trajectory RMSE of \textbf{0.3492 m} and an endpoint success rate of \textbf{74.40\%} on the 1,000-sample DiffWAM-1000 benchmark, while representative real-world experiments demonstrate complex behaviors including constrained traversal, orbiting, S-shaped flight, and multi-stage navigation. An onboard DiffWAM-Flash implementation further reaches \textbf{1.08 s} model-pipeline latency on NVIDIA Jetson AGX Thor. These results demonstrate that predictive video representations can be efficiently grounded into continuous 3D motion, providing a direct alternative to generate-then-reconstruct navigation pipelines. \textcolor{blue}{Project page: \url{https://zzmmzzm.github.io/diffwam.github.io/}}.
% 预训练视频基础模型蕴含丰富的语义与时空先验，但将这些先验转化为无人机运动通常需要代价高昂的未来视频生成和几何重建。本文研究能否直接从冻结视频模型的预测表征中恢复未来视觉预测所隐含的运动。为此，我们提出 \textbf{DiffWAM}，一种几何条件导航世界动作模型，可将多层预测特征直接映射为连续相机轨迹。其中，\textbf{Grid-Motion} 保留空间与时间维度上的运动关联，\textbf{Latent2Pose} 结合首帧几何信息，将预测表征进一步映射为具有米制尺度的三维运动。完整视频 rollout 与几何重建仅用于离线监督，因此部署阶段无需未来视频解码和在线多帧几何重建。进一步地，我们提出 \textbf{FastDreamer}，在无人机持续飞行过程中并行执行预测与几何计算，并通过带时间戳的异步轨迹交接实现连续执行。
% 在包含 1,000 个样本的 DiffWAM-1000 基准上，DiffWAM 实现了 \textbf{0.3492 m} 的轨迹 RMSE 和 \textbf{74.40\%} 的端点成功率；真实无人机实验进一步展示了受限空间穿越、环绕、S 型飞行以及多阶段导航等复杂运动能力。端侧 DiffWAM-Flash 实现可在 NVIDIA Jetson AGX Thor 上达到 \textbf{1.08 s} 的模型流水线时延。上述结果表明，视频模型的预测表征可以被高效地映射为连续三维运动，为传统“生成视频--几何重建”式导航流程提供了一条更加直接的技术路径。

\end{abstract}
\keywords{World-action models, predictive video representations, trajectory grounding, geometric distillation, 3D navigation}
% 中文翻译：关键词：世界动作模型、预测视频表征、轨迹grounding、几何蒸馏、无人机导航。

% ============================================================================
% 本章内容：从生成视频到可执行运动的接口问题出发，提出冻结预测表征的几何grounding，说明训练与执行设计，并用已有证据界定贡献。
% ============================================================================
\section{Introduction}
\label{sec:introduction}

With the rapid advancement of large language models and multimodal foundation models, 3D vision-language navigation/action (VLN/VLA) ~\cite{zhang2026embodied,wu2025vla,indooruav,uavflow}, particularly for unmanned aerial vehicles (UAVs), has become an important research direction in robotics. Existing methods can combine target recognition with instruction understanding to accomplish object-goal navigation tasks such as ``navigate to the electric fan next to the table.'' However, they remain less effective for instructions involving continuous, structured motion, such as ``circle around the fire hydrant three times for inspection.'' A key limitation is that current VLN/VLA systems based on vision language models (VLMs) ~\cite{bai2023qwenvl} are generally better at deciding where to go or what to do next than at translating complex language instructions into continuous, executable trajectories subject to direction, clearance, and spatial-scale constraints. In contrast, video foundation models naturally learn how visual scenes evolve over time, making their internal representations better aligned with continuous motion processes.
% 中文翻译：
% 随着大语言模型和多模态基础模型的快速发展，三维视觉语言导航（Vision-Language Navigation, VLN），尤其是面向无人机（UAV）的视觉语言导航，已成为机器人领域的重要研究方向。现有方法已经能够结合目标识别与指令理解，完成诸如“导航到桌子旁边的电风扇”这样的目标导向导航任务。然而，对于“环绕消防栓三圈进行检查”这类包含连续、结构化运动的复杂指令，现有方法的表现仍然有限。其核心原因在于，当前基于VLM的 VLN/VLA 系统更擅长判断“去哪里”或“下一步做什么”，但较难进一步将复杂语言指令转化为满足飞行方向、安全间隙和空间尺度约束的连续可执行轨迹。相比之下，视频基础模型天然学习视觉场景随时间的连续演化过程，因此其内部表征与连续运动过程具有更强的一致性。

In our previous work, NavDreamer~\cite{navdreamer}, we introduced a video-foundation-model-based \textbf{generate--reconstruct} paradigm: first generating a future video and then recovering the implicit camera trajectory from it. This demonstrated that video foundation models can provide rich semantic and temporal priors for 3D navigation. However, complete video generation and pixel-level rendering introduce substantial computation that is not intrinsically required for navigation. In practice, image fidelity, texture detail, and color realism are far less important than the underlying sequence of motion states and camera poses. This motivates a more direct question: \textit{before the future video is fully generated, do the intermediate representations of a video generative model already contain sufficient information to recover the underlying navigation motion?}
% 中文翻译：
% 在我们的前期工作中，我们提出了一种基于视频基础模型的 \textbf{generate--reconstruct} 范式名为NavDreamer：首先生成未来视频，再从中恢复其隐含的相机运动轨迹。该方法表明，视频基础模型能够为三维导航提供丰富的语义与时序先验。然而，完整视频生成和像素级渲染会引入大量并非导航本质所必需的计算开销。对于导航而言，图像清晰度、纹理细节和颜色真实性远不如其背后对应的连续运动状态和相机位姿序列重要。因此，我们进一步提出一个更直接的问题：\textbf{在未来视频尚未完整生成之前，视频生成模型的中间表征是否已经包含足以恢复其潜在导航运动的信息？}

Recent implicit World-Action Models (WAMs) \cite{fastwam} provide a promising direction by directly decoding actions from video-model representations. However, existing approaches often aggregate features from many hidden layers, leading to unnecessarily large action branches, while early predictive features themselves are not explicit pose representations: they entangle appearance, camera motion, object motion, language conditions, and prediction uncertainty. Naive global pooling may therefore destroy the spatial correspondences required for motion recovery. Moreover, representations at different network depths contribute differently to navigation, and implicit features alone do not provide a reliable metric spatial reference. These observations lead to three key research questions: \textit{(1) At which denoising stage does a video foundation model begin to form navigation-motion representations that can be decoded reliably? (2) Which network depths and latent representations are most informative for action generation? (3) How can predictive visual knowledge be efficiently grounded into continuous, executable trajectories with geometric, metric-scale, and temporal constraints?}
% 中文翻译：
% 近年来，隐式世界动作模型（World-Action Models, WAMs）提供了一种有前景的方向，即直接从视频模型表征中解码动作。然而，现有方法通常聚合大量隐藏层特征，导致动作分支规模过大；与此同时，早期预测特征本身并不是显式位姿表征，而是混合了场景外观、相机运动、物体运动、语言条件以及未来预测的不确定性。因此，直接进行全局池化可能破坏运动恢复所依赖的空间对应关系。此外，不同网络深度的表征对导航运动的贡献并不相同，而仅依赖隐式表征本身也难以获得可靠的米制空间参考。基于这些观察，我们提出三个关键研究问题：\textit{(1) 视频基础模型在去噪过程的哪个阶段开始形成能够被可靠解码的导航运动表征？(2) 哪些网络深度和潜在表征对于动作生成最具信息量？(3) 如何将预测视觉知识高效地映射为满足几何、米制尺度和时序约束的连续可执行轨迹？}

To answer these questions, we propose \textbf{DiffWAM}, a navigation World Action Model that directly transforms the predictive representations of a pretrained video foundation model into continuous 3D motion. Our design follows the three questions above. First, rather than waiting for complete video synthesis, we investigate whether navigation motion can already be decoded from an early predictive state of the frozen video model, allowing subsequent generation to be bypassed once sufficiently informative representations have been obtained. Second, because different network depths encode complementary visual, semantic, and motion information, we selectively read and fuse representations from multiple depths instead of forwarding a large collection of hidden layers to the action branch. Third, since these implicit representations do not by themselves define a metrically meaningful trajectory, we preserve their spatial-temporal structure before pooling and condition motion decoding on geometry extracted from the initial observation, thereby grounding predictive visual knowledge into a common spatial reference and estimated metric scale. We instantiate these operations with Grid-Motion, which establishes motion associations across spatial locations and time, and Latent2Pose, which converts the resulting motion representation into a time-indexed sequence of camera poses.
% 中文翻译：
% 为回答上述三个问题，我们提出 \textbf{DiffWAM}，一种将预训练视频基础模型的预测表征直接转换为连续三维运动的导航世界动作模型。其设计与前述三个问题逐一对应。首先，我们不再等待完整未来视频生成，而是研究能否从冻结视频模型的早期预测状态中直接解码导航运动，使系统在获得足够有效的预测表征后即可绕过后续生成过程。其次，考虑到不同网络深度分别包含互补的视觉、语义与运动信息，我们不再将大量隐藏层统一送入动作分支，而是有选择地读取并融合多个深度的预测特征。最后，由于这些隐式表征本身并不能直接定义具有米制意义的轨迹，我们在池化前保留其时空结构，并利用初始观测中提取的几何信息对运动解码进行条件约束，从而将预测视觉知识映射到统一的空间参考和估计的米制尺度中。具体而言，我们通过 Grid-Motion 建立跨空间位置与时间的运动关联，并利用 Latent2Pose 将得到的运动表征进一步转换为带时间索引的相机位姿序列。

Learning such a direct predictive-to-motion mapping, however, requires supervision that reveals the motion implicitly represented by these early predictive features. We therefore adopt an asymmetric teacher--student training strategy that exploits complete video generation only during offline training. Starting from the same predictive state observed by the student, the teacher continues the video-generation process and reconstructs the resulting future camera motion, with geometric calibration providing a metric trajectory reference. The student is then trained to recover this trajectory using only the paired early predictive representations and first-frame geometry. In this way, motion information that would otherwise require expensive future generation and reconstruction to obtain explicitly is distilled into a compact trajectory readout, while the video backbone and geometry estimator remain frozen. At deployment, complete video rollout, video decoding, and multi-frame geometric reconstruction are entirely removed. Experiments confirm that this direct readout preserves useful navigation motion: DiffWAM achieves a trajectory RMSE of 0.3492 m and an endpoint success rate of 74.40\% on DiffWAM-1000, while real-world flights further demonstrate constrained traversal, orbiting, S-shaped motion, landing, and multi-stage navigation.
% 中文翻译：
% 然而，要学习这种从预测表征直接映射到运动的关系，还需要能够揭示这些早期预测特征中隐含运动信息的监督信号。为此，我们采用一种非对称 teacher--student 训练策略，仅在离线训练阶段使用完整视频生成。从学生所观测到的同一预测状态出发，教师继续执行后续视频生成过程，并通过几何重建恢复未来相机运动，同时利用几何标定获得具有米制尺度的参考轨迹。学生则仅使用与该完整生成过程配对的早期预测表征和首帧几何信息，学习恢复对应的运动轨迹。通过这种方式，原本需要经过昂贵未来生成和几何重建才能显式获得的运动信息，被蒸馏到紧凑的轨迹解码器中，而视频骨干网络和几何估计器始终保持冻结。部署阶段则完全移除完整视频 rollout、视频解码以及多帧几何重建。实验结果表明，这种直接运动解码能够有效保留导航所需的信息：在受控轨迹对比中，DiffWAM 实现了 0.3492 m 的轨迹 RMSE，并在 DiffWAM-1000 上取得 74.40\% 的端点成功率；真实无人机实验进一步展示了受限空间穿越、环绕、S 型运动、降落以及多阶段导航等复杂运动能力。

\begin{figure}[!t]
    \centering
    \includegraphics[width=\linewidth]{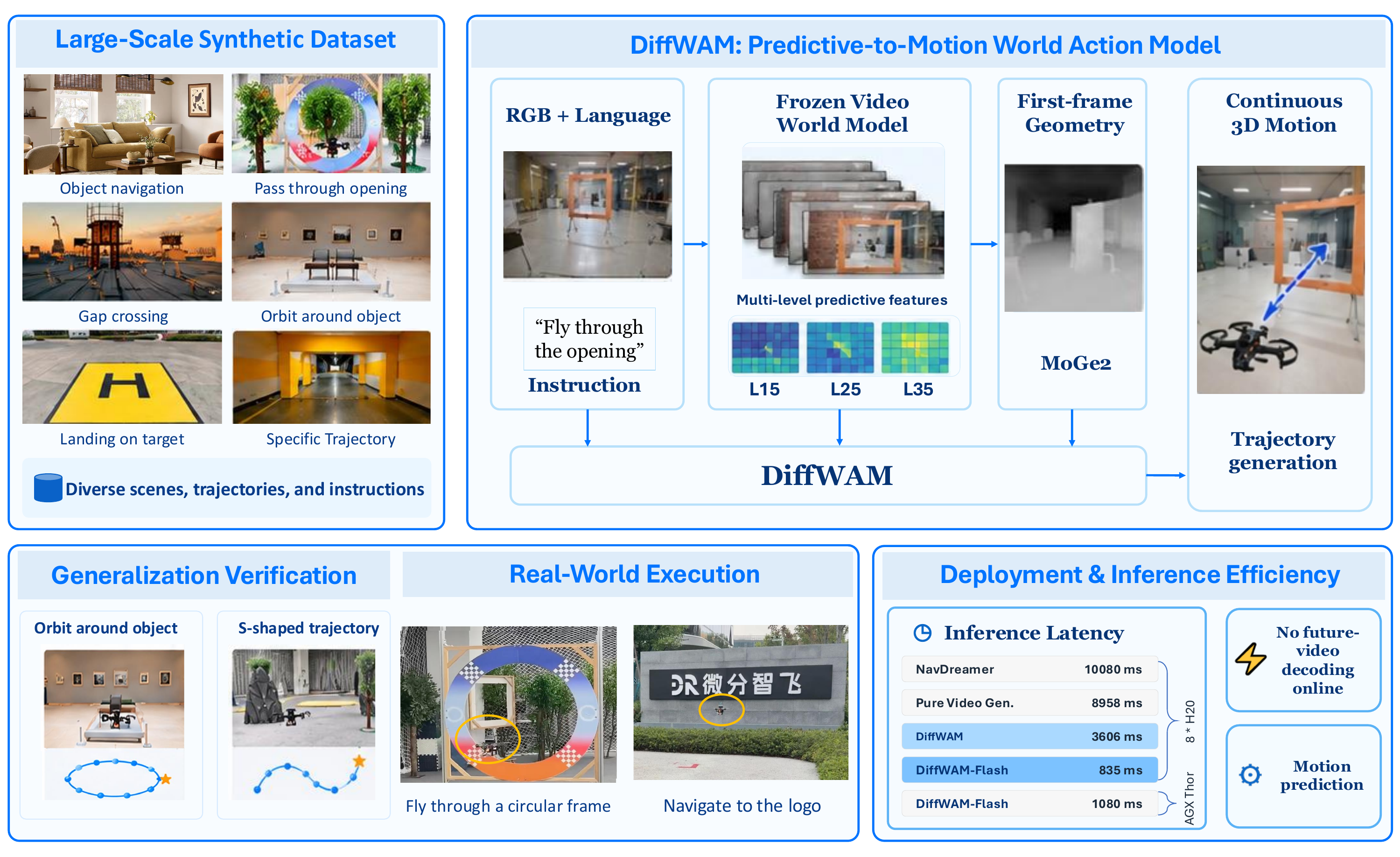}
\caption{\textbf{Overview of DiffWAM.} DiffWAM grounds multi-level predictive representations from a frozen video world model, together with first-frame geometry, into continuous 3D UAV trajectories. Trained with large-scale synthetic navigation data, it supports diverse structured motions, real-world execution, and efficient deployment without online future-video decoding.}
% 中文翻译：DiffWAM 总体流程。DiffWAM 通过 Grid-Motion 和 Latent2Pose，将所选早期预测特征与首帧几何转换为相机运动轨迹。训练依次包括视频骨干准备（S0）、导航感知读出预训练（S1）以及几何引导的预测蒸馏（S2）；其中，完整视频 rollout 及其几何重建为蒸馏阶段提供离线轨迹监督。详细张量和输出维度见 Fig.~\ref{fig:DiffWAM_Internal_model_architecture}。
    \label{fig:DiffWAM_toutu}
\end{figure}

Direct trajectory decoding substantially shortens the computational path from visual prediction to motion generation, but efficient prediction alone does not guarantee continuous flight. During onboard inference, the UAV continues to move while a new trajectory is being computed; consequently, when a proposal becomes available, the vehicle state may already have changed and the proposal may no longer be aligned with the state and reference frame from which it was predicted. We therefore further introduce \textbf{FastDreamer}, an inference-and-execution framework that extends DiffWAM from individual motion prediction to continuous closed-loop navigation. FastDreamer reduces proposal-preparation overhead by reusing the world-model conditioning components for local instruction rewriting and by executing predictive-representation extraction and first-frame geometry estimation in parallel. More importantly, computation is overlapped with execution: while the UAV follows its currently committed trajectory, the next motion proposal is prepared in advance. Each update retains its observation timestamp and is associated with a scheduled future handoff state, allowing the downstream planner to explicitly account for the vehicle motion accumulated during inference and construct a compatible transition before activation. This design addresses both the latency of preparing a new motion proposal and the state/reference mismatch introduced while the UAV continues to move, extending predictive-to-motion grounding toward continuous and multi-stage vision-language UAV navigation.
% 中文翻译：
% 直接轨迹解码显著缩短了从视觉预测到运动生成的计算链路，但仅仅提高单次预测效率仍不足以保证连续飞行。在端侧推理过程中，无人机在计算新轨迹的同时仍会持续运动，因此当新的运动提议计算完成时，飞行器状态可能已经发生变化，使该轨迹与其最初预测时对应的状态和参考系产生偏移。为此，我们进一步提出 \textbf{FastDreamer}，一种将 DiffWAM 从单次运动预测扩展到连续闭环导航的推理与执行框架。FastDreamer 一方面通过复用世界模型中的条件编码组件完成局部指令重写，并并行执行预测表征提取和首帧几何估计，以降低轨迹准备开销；更重要的是，它将计算过程与当前轨迹执行进行重叠：当无人机持续跟踪已经提交的轨迹时，系统同步提前计算下一段运动提议。每次更新均保留其观测时间戳，并关联一个预定的未来交接状态，使下游规划器能够显式考虑推理过程中累积的飞行器运动，并在激活新轨迹之前构造与当前执行状态兼容的过渡。由此，该设计同时解决了下一段运动提议的计算延迟，以及无人机在推理期间持续运动所引入的状态与参考系偏移，从而将预测表征到运动的映射进一步扩展到连续、多阶段的视觉语言无人机导航。

Our contributions are fourfold:
% 中文翻译：本文的主要贡献包括以下四个方面：
\begin{enumerate}

    \item \textbf{A direct predictive-to-motion formulation for navigation WAMs.}
    We formulate continuous UAV navigation as directly decoding camera motion from intermediate predictive representations of a pretrained video foundation model, avoiding complete future-video synthesis at deployment.
    % 中文翻译：
    % \textbf{面向导航WAM的直接预测表征到运动建模。}
    % 我们将连续无人机导航建模为直接从预训练视频基础模型的中间预测表征中解码相机运动，从而在部署阶段避免完整未来视频生成。

    \item \textbf{Geometry-grounded motion decoding and predictive distillation.}
    We develop a geometry-conditioned motion readout that preserves spatial-temporal structure and learns from rollout-paired offline supervision, enabling camera-trajectory prediction while keeping the video and geometry backbones frozen.
    % 中文翻译：
    % \textbf{几何约束的运动解码与预测蒸馏。}
    % 我们设计了几何条件化的运动读出模块，在保留时空结构的同时利用同rollout离线监督进行学习，从而在冻结视频与几何骨干网络的条件下实现米制相机轨迹预测。

    \item \textbf{Latency-aware continuous execution with FastDreamer.}
    We introduce FastDreamer to bridge trajectory prediction and continuous UAV execution through parallel inference, flight-time-aware scheduling, and timestamp-aware prospective handoff.
    % 中文翻译：
    % \textbf{基于FastDreamer的延迟感知连续执行。}
    % 我们提出FastDreamer，通过并行推理、面向飞行时间的调度以及时间戳感知的前瞻式轨迹交接，将轨迹预测进一步扩展到连续无人机执行。

    \item \textbf{Comprehensive validation across navigation settings.}
    We evaluate DiffWAM-FastDreamer across benchmark, simulation, real-world flight, onboard deployment, and controlled ablation studies, demonstrating its effectiveness across diverse continuous and structured UAV navigation tasks.
    % 中文翻译：
    % \textbf{面向多种导航场景的系统验证。}
    % 我们通过基准测试、仿真、真实飞行、端侧部署及系统消融实验对DiffWAM进行评估，验证其在多种连续与结构化无人机导航任务中的有效性。

\end{enumerate}

% ============================================================================
% 本章内容：按预测表征与动作接口、几何监督与连续执行两条线索比较既有方法，分析各类方法解决的变量、适用条件及本文的具体定位。
% ============================================================================
\section{Related Work}
\label{sec:related_work}

\subsection{Video World Models for Embodied Action and Navigation}

Vision--language--action models map multimodal observations and instructions to robot actions by adapting pretrained vision--language representations to embodied control. OpenVLA~\cite{kim2024openvla} demonstrates this paradigm at scale, while some work~\cite{chen2026towards} extends it to efficient onboard aerial navigation. In parallel, video foundation models provide predictive priors that capture scene evolution and motion. DreamZero~\cite{ye2026world} and WorldFly~\cite{worldfly} jointly model future visual states and actions, showing that video prediction can support embodied decision making.

For aerial navigation, NavDreamer~\cite{navdreamer} and ImagineUAV~\cite{imagineuav} generate future visual observations and subsequently recover 3D motion, whereas WorldVLN~\cite{worldvln} predicts latent world transitions and decodes waypoint actions. Recent Fast-WAM~\cite{fastwam} and Faster-WAM~\cite{fasterwam} further show that useful action prediction does not necessarily require complete future rendering at test time. DiffWAM follows this direction but focuses on a different interface: it directly grounds intermediate predictive representations of a frozen video model into continuous camera trajectories, removing complete future-video decoding from deployment.

\subsection{Predictive Representation Grounding and Distillation}

Predictive video features encode motion together with appearance, semantics, and uncertainty, but do not directly provide a metric 3D trajectory. Geometric foundation models offer complementary spatial information. $\pi^3$~\cite{pi3} reconstructs camera motion and scene geometry from image collections, while MoGe-2~\cite{wang2025moge2} estimates metric geometry from a single image. DiffWAM uses first-frame geometry to provide spatial reference and scale during deployment, while complete-video reconstruction is used only to construct offline trajectory supervision.

This training setting is related to knowledge and policy distillation. Conventional knowledge distillation transfers~\cite{cui2026revisiting} predictive or representational information between teacher and student models~\cite{cui2025multilevelot,11672289}, while policy distillation transfers behavior from a teacher policy~\cite{rusu2015policy}. DiffWAM instead distills the motion implied by an expensive future continuation and reconstruction process. The student observes only early predictive representations and first-frame geometry, while the teacher trajectory is reconstructed from the corresponding completed video rollout. The resulting supervision therefore transfers future predictive information into a direct representation-to-trajectory readout rather than matching teacher logits or action distributions.

\subsection{Latency-Aware Continuous Execution}

Foundation-model inference can be slow relative to robotic control, motivating both model-side acceleration and asynchronous execution. Speculative inference provides one route to reducing large-model computation, with recent work characterizing its scaling behavior across model and inference configurations~\cite{yan2025scaling}. At the policy level, Diffusion Policy~\cite{chi2023diffusionpolicy} adopts receding-horizon execution, while Real-Time Chunking~\cite{black2025rtc} overlaps action generation with ongoing execution. AsyncVLA~\cite{hirose2026asyncvla} similarly separates slower semantic reasoning from faster onboard control.

For mobile robots, delay also introduces spatial inconsistency because the robot moves between observation and action activation. For instance, PathPainter~\cite{wang2026pathpainter} has constructed a global planner and a local planner based on API, and during execution, they need to be executed asynchronously to alleviate the latency. AsyncShield~\cite{yang2026asyncshield} compensates for delayed navigation outputs using pose-aware geometric alignment, while classical UAV planners such as EGO-Planner~\cite{ego_planner} provide dynamically feasible local trajectory generation. FastDreamer addresses the complementary interface between delayed predictive motion and continuous UAV execution by preparing new proposals during ongoing flight and associating them with their observation time and scheduled handoff state.

\section[DiffWAM: Grounding Predictive Representations into Motion]{DiffWAM: Grounding Predictive Representations into Motion}
% 中文翻译：DiffWAM：将预测表征映射为运动
\label{sec:method}

As motivated in Sec.~\ref{sec:introduction}, our objective is to recover language-conditioned navigation motion from the intermediate predictive representations of a video world model without completing expensive video synthesis. This requires retaining motion-relevant spatiotemporal information and grounding it in a meaningful spatial reference and estimated metric scale. We develop DiffWAM as a direct representation-to-motion interface that produces camera-trajectory proposals for continuous 3D navigation. In this work, we instantiate DiffWAM with MiniMax H3\cite{minimaxh3} as the video world model backbone and read out its intermediate predictive representations for motion grounding.
% 中文翻译：如引言所述，我们的目标是在无需完成昂贵视频合成的情况下，从视频世界模型的中间预测表征中恢复语言条件下的导航运动。这需要保留与运动相关的时空信息，并将其映射到明确的空间参考和估计米制尺度。为此，我们提出 DiffWAM，作为直接的表征到运动接口，为连续三维导航生成相机轨迹提议。本文中，我们采用 MiniMax H3 作为 DiffWAM 的视频世界模型骨干，并读取其中间预测表征以进行运动映射。

\begin{figure}[!h]
    \centering
    \includegraphics[width=\linewidth]{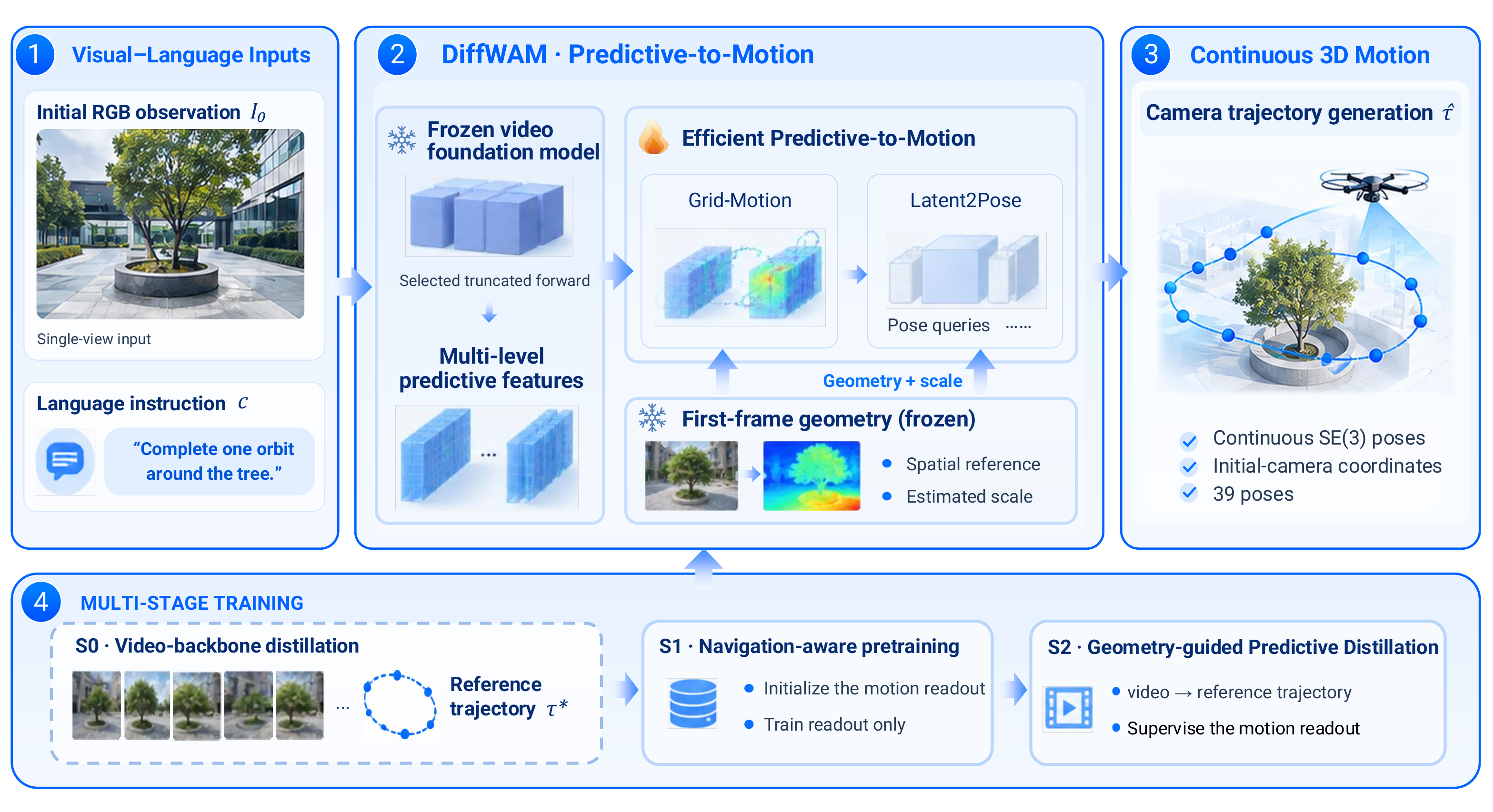}
    \vspace{-0.5cm}
    \caption{\textbf{DiffWAM's pipeline.} DiffWAM converts selected early predictive features and first-frame geometry into camera-motion trajectories through Grid-Motion and Latent2Pose. Training proceeds through video-backbone preparation (S0), navigation-aware readout pretraining (S1), and geometry-guided predictive distillation (S2), where completed video rollouts and geometric reconstruction provide offline trajectory supervision. Detailed tensor and output dimensions are specified in Fig.~\ref{fig:DiffWAM_Internal_model_architecture}.}
% 中文翻译：DiffWAM 总体流程。DiffWAM 通过 Grid-Motion 和 Latent2Pose，将所选早期预测特征与首帧几何转换为相机运动轨迹。训练依次包括视频骨干准备（S0）、导航感知读出预训练（S1）以及几何引导的预测蒸馏（S2）；其中，完整视频 rollout 及其几何重建为蒸馏阶段提供离线轨迹监督。详细张量和输出维度见 Fig.~\ref{fig:DiffWAM_Internal_model_architecture}。
    \label{fig:DiffWAM_Pipeline}
\end{figure}

Given an initial RGB observation and a language instruction, DiffWAM extracts multi-level predictive features using four successive evaluations of the frozen video backbone, retaining features at zero-based schedule indices 0-3. The first three evaluations are completed to advance the video latent, while the fourth terminates immediately after DiT block 35. DiffWAM-Flash retains features only at schedule index 0 and terminates its first evaluation after the same block. Unless otherwise stated, DiffWAM denotes the four-evaluation configuration. Grid-Motion preserves spatial correspondences and establishes geometry-conditioned motion associations, while Latent2Pose combines predictive representations with first-frame geometry estimated by frozen MoGe2 to recover a time-indexed sequence of camera poses. Neither variant performs future-video VAE decoding or multi-frame reconstruction at deployment.
% 中文翻译：给定初始 RGB 观测和语言指令，DiffWAM 连续执行四次冻结视频骨干前向，并保留从 0 开始编号的调度状态 0、1、3 的多层预测特征。前三次前向完整执行以更新视频 latent，第四次在 DiT 第 35 号块之后立即退出。DiffWAM-Flash 仅保留调度状态 0 的特征，在第一次前向的同一块之后退出。除非另有说明，DiffWAM 表示四次前向版本。Grid-Motion 保留空间对应关系并建立几何条件下的运动关联，Latent2Pose 将预测表征与冻结的 MoGe2 所估计的首帧几何结合，恢复带时间索引的相机位姿序列。两种版本在部署时均不进行未来视频 VAE 解码或多帧重建。
Training follows a three-stage procedure, as illustrated in Fig.~\ref{fig:DiffWAM_Pipeline}: video-backbone distillation (S0), navigation-aware motion-readout pretraining (S1), and geometry-guided predictive distillation (GPD, S2). Completed video rollouts and geometric reconstruction provide offline trajectory supervision. GPD pairs early predictive features with reference trajectories reconstructed from the same video rollout, transferring trajectory-relevant information from completed video predictions to the motion readout. The video backbone and the MoGe2 geometry estimator remain frozen during motion-readout training; the trainable scope of the readout is specified below. This asymmetric training procedure removes complete future-video synthesis and subsequent trajectory reconstruction from deployment.
% 中文翻译：如总体流程图所示，训练包括三个阶段：视频骨干准备 S0、导航感知运动读出预训练 S1，以及几何引导的预测蒸馏 GPD（S2）。完整视频 rollout 及其几何重建提供离线轨迹监督。GPD 将早期预测特征与同一次视频 rollout 重建得到的参考轨迹配对，把完整视频预测中的轨迹信息迁移到运动读出器。运动读出训练期间，视频骨干与 MoGe2 几何估计器保持冻结；读出器的具体可训练范围在下文说明。这种非对称训练方式使部署阶段无需完整生成未来视频，也无需随后重建轨迹。

% =============================================================================
\subsection{Problem Formulation}
% 中文翻译：问题定义
\label{sec:formulation}

\begin{figure}[!h]
    \centering
    \includegraphics[width=\linewidth]{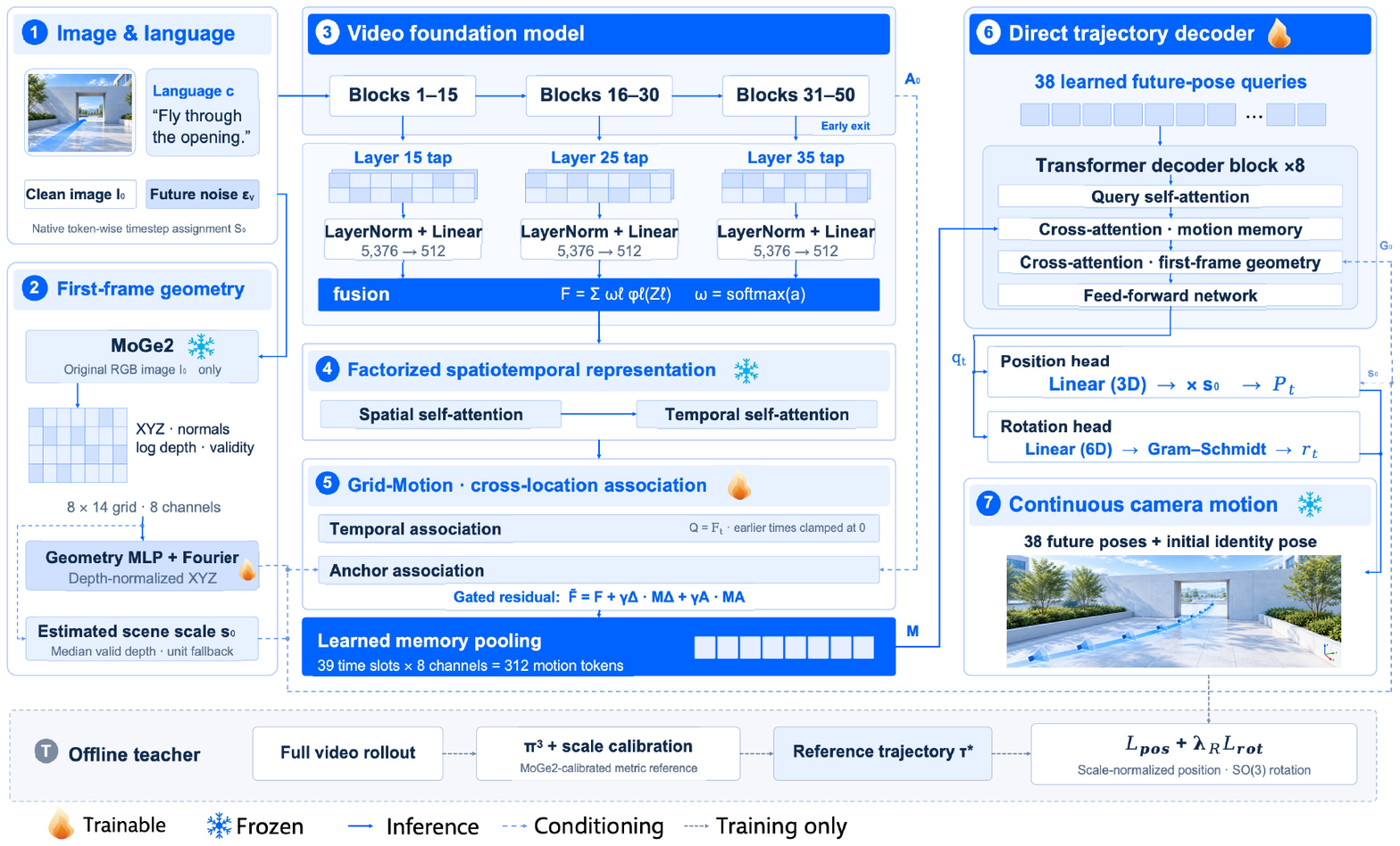}
    \caption{\textbf{DiffWAM's model architecture.}
    Features from DiT blocks 15, 25, and 35 are projected to 512 dimensions, fused, and processed by a frozen factorized spatiotemporal module. Grid-Motion associates predictive features across time and with geometry-conditioned anchors before pooling them into 312 motion-memory tokens. A direct decoder with 38 pose queries and eight transformer blocks predicts 38 future poses from motion memory and first-frame geometry; together with the initial identity pose, these form a 39-pose trajectory. During offline supervision, $\pi^3$ reconstructs camera motion and depth from completed video rollouts, and MoGe2 provides metric-depth information for translation-scale calibration.}
    % 中文翻译：DiffWAM 内部架构。来自 DiT 第 15、25 和 35 个块的特征首先投影至 512 维并进行融合，随后由冻结的分解式时空模块处理。Grid-Motion 在池化之前建立跨时间的预测特征关联以及与几何条件锚点之间的关联，最终形成 312 个运动记忆 token。直接轨迹解码器包含 38 个位姿 query 和 8 个 Transformer 模块，根据运动记忆和首帧几何预测 38 个未来位姿；再与初始单位位姿结合，形成包含 39 个位姿的轨迹。在离线监督阶段，$\pi^3$ 从完整视频 rollout 中重建相机运动与深度，而 MoGe2 提供米制深度信息，用于平移尺度校准。
    \label{fig:DiffWAM_Internal_model_architecture}
\end{figure}

Given an initial RGB observation $I_0$ and a language instruction $c$, DiffWAM predicts camera motion over a nominal five-second horizon. The future trajectory is represented as
\begin{equation}
\hat{\tau}
=
\{(\hat{R}_t,\hat{p}_t)\}_{t=1}^{T},
\qquad T=38,
\label{eq:trajectory_definition}
\end{equation}
where $\hat{R}_t\in\mathrm{SO}(3)$ and $\hat{p}_t\in\mathbb{R}^{3}$ denote the orientation and position at the $t$-th future timestamp. We express all future camera poses relative to the initial camera frame: a point $x^{C_t}$ in the future camera frame is mapped to the initial camera frame as $x^{C_0}=\hat{R}_t x^{C_t}+\hat{p}_t$. The initial camera axes point right, down, and forward. Prepending $(R_0,p_0)=(I_3,\mathbf{0})$ gives 39 poses in total, as specified by the detailed architecture in Fig.~\ref{fig:DiffWAM_Internal_model_architecture}.
% 中文翻译：其中，$\hat{R}_t\in\mathrm{SO}(3)$ 和 $\hat{p}_t\in\mathbb{R}^{3}$ 分别表示第 $t$ 个未来时间戳的姿态和平移。所有未来相机位姿均相对于初始相机坐标系表示：未来相机坐标系中的点 $x^{C_t}$ 通过 $x^{C_0}=\hat{R}_t x^{C_t}+\hat{p}_t$ 映射到初始相机坐标系。初始相机坐标轴分别指向右、下和前方。加入初始单位位姿 $(R_0,p_0)=(I_3,\mathbf{0})$ 后，总共得到 39 个位姿，与 Fig.~\ref{fig:DiffWAM_Internal_model_architecture} 中的详细架构一致。

Each pose query is associated with a prescribed time offset $\delta_t$, with $0=\delta_0<\delta_1<\cdots<\delta_T$, shared by prediction and supervision. These offsets index physical video time rather than denoising steps. Here, continuous-valued poses distinguish the output from discrete action tokens; a finite pose sequence does not by itself specify a continuously differentiable or dynamically feasible flight trajectory. Temporal interpolation and executable trajectory construction belong to the downstream execution system.
% 中文翻译：每个位姿 query 对应一个预定时间偏移，预测与监督共享同一组时间戳。这些时间偏移表示视频中的物理时间，而不是去噪步数。连续值位姿与离散动作 token 相区别，但有限位姿序列本身不等于连续可微或满足动力学约束的飞行轨迹。时间插值及可执行轨迹构建由下游执行系统完成。

Rather than completing the video-generation schedule, DiffWAM retains predictive features from selected early backbone evaluations. Let $\mathcal{K}$ denote the retained zero-based schedule indices, with $\mathcal{K}=\{0,1,2,3\}$ for DiffWAM and $\mathcal{K}=\{0\}$ for DiffWAM-Flash. We denote the retained feature collection by $Z=\{Z^{(k)}\}_{k\in\mathcal{K}}$ and write
\begin{equation}
(Z,A_0)
=
\mathcal{E}_{\psi}^{\mathcal{K}}(I_0,c,\epsilon_v),
\label{eq:h3_predictive_state}
\end{equation}
where $\epsilon_v$ is the sampled video noise, $\psi$ denotes the frozen video-model parameters, and $A_0$ denotes the observation-anchor representation supplied to the readout. The extraction operator includes all preceding backbone evaluations and scheduler updates needed to reach $\max\mathcal{K}$, and terminates after DiT block 35 of the final required evaluation. The frozen MoGe2 geometry estimator provides
\begin{equation}
G_0=\mathcal{G}_{\eta}(I_0),
\label{eq:first_frame_geometry}
\end{equation}
and the motion-readout interface predicts
\begin{equation}
\hat{\tau}=\mathcal{D}_{\theta}(Z,A_0,G_0).
\label{eq:latent2pose}
\end{equation}
The parameter set $\theta$ contains the trainable readout components and excludes the frozen video backbone and MoGe2 estimator.
% 中文翻译：DiffWAM 不完成视频生成调度，而是保留所选早期骨干前向中的预测特征。K 表示从 0 开始编号的所选调度状态：DiffWAM 使用 {0,1,2,3}，DiffWAM-Flash 使用 {0}。Z 表示这些状态的特征集合，A_0 表示提供给读出器的观测锚点表征。特征提取算子包含到达最后一个所选状态所需的全部先行骨干前向和调度器更新，并在最后一次所需前向的 DiT 第 35 号块后终止。冻结的 MoGe2 从初始图像提取几何，运动读出器结合预测特征、观测锚点与首帧几何预测轨迹。可训练参数 theta 不包含视频骨干和 MoGe2 的冻结参数。

Each retained schedule state preserves the native token-wise timestep assignments. A schedule index identifies a denoising evaluation, whereas a DiT block index identifies network depth within that evaluation; both differ from the physical timestamps of the predicted trajectory. The four-evaluation and one-evaluation budgets refer exclusively to the video backbone, not to the internal sampling budget of the pose readout. Neither variant performs future-video VAE decoding.
% 中文翻译：每个保留的调度状态均使用原生逐 token 时间步分配。调度索引表示去噪前向的次序，DiT 块索引表示该次前向内部的网络深度；两者都不同于预测轨迹的物理时间戳。四次和一次前向预算仅指视频骨干，不代表位姿读出器内部的采样次数。两种版本均不进行未来视频 VAE 解码。
Features from states $\mathcal{K}=\{0,1,2,3\}$ are spatially encoded with state, position, and noise-level embeddings, concatenated at each video slot, and fused by learned attention pooling:
\begin{equation}
m_u=\operatorname{MHA}\!\left(q_{\mathrm{frame}},
\operatorname{Concat}_{k\in\mathcal{K}}H_u^{(k)},
\operatorname{Concat}_{k\in\mathcal{K}}H_u^{(k)}\right).
\end{equation}
% 中文翻译：状态 {0,1,2,3} 的特征经空间编码并加入状态、位置和噪声水平嵌入后，在每个视频槽内拼接，通过可学习注意力池化融合。

% =============================================================================
\subsection{Efficient Predictive Motion Readout}
% 中文翻译：高效预测运动读出
\label{sec:predictive_features}

DiffWAM investigates whether an early computation of a trained video model contains motion information that can be recovered without decoding future RGB frames. The readout therefore operates on hidden features of the video backbone rather than on rendered video or reconstructed future geometry.
% 中文翻译：DiffWAM 研究经过训练的视频模型在早期计算中是否包含无需解码未来 RGB 帧即可读取的运动信息。因此，读出过程直接作用于 H3 隐藏特征，而不是渲染后的视频或未来重建几何。

\paragraph{Multi-level predictive representation.}
For each retained schedule index $k\in\mathcal{K}$, we extract hidden features from DiT blocks 15, 25, and 35 using zero-based block indexing. The three depth taps are collected within the same backbone evaluation, whereas different schedule indices correspond to successive denoising evaluations. Only the final required evaluation is truncated after block 35. Before readout-specific feature preparation, the per-evaluation tensors are represented as
\begin{equation}
Z^{(k)}\in\mathbb{R}^{L\times T_v\times H_z\times W_z\times C},
\qquad
A_0^{(k)}\in\mathbb{R}^{L\times 1\times H_z\times W_z\times C},
\label{eq:feature_shape}
\end{equation}
where $L=3$, $T_v=39$ counts predictive video-time slots, $H_z\times W_z$ is the spatial feature grid, and $C$ is the hidden-channel dimension. The retained predictive input is $Z=\{Z^{(k)}\}_{k\in\mathcal{K}}$. The video-time dimension, denoising schedule index, and DiT block index describe three distinct axes and must not be conflated. Observation-anchor features do not introduce additional output poses.
% 中文翻译：对于每个所选调度索引 k，我们提取从 0 开始编号的 DiT 第 15、25、35 号块隐藏特征。同一状态的三个深度特征来自同一次骨干前向，不同调度状态则对应连续的去噪前向。只有最后一次所需前向在第 35 号块后截断。公式描述读出器特定特征整理之前的单次前向张量：L=3 表示所选深度数，T_v 表示预测视频时间槽数，H_z×W_z 表示空间特征网格，C 表示隐藏通道数。读出输入 Z 汇集所有所选状态的预测特征。视频时间维、去噪调度索引与 DiT 块索引是三个不同维度，不能混用。观测锚点特征不会增加输出位姿数量。

Each depth passes through a layer-specific normalization and a projection from 5,376 to 512 channels. The projected features are fused using softmax-normalized coefficients:
\begin{equation}
F^{\mathrm{fuse}}
=
\sum_{\ell=1}^{L}\omega_{\ell}\phi_{\ell}(Z_{\ell}),
\qquad
\omega_{\ell}
=
\frac{\exp(a_{\ell})}{\sum_{j=1}^{L}\exp(a_j)}.
\label{eq:layer_fusion}
\end{equation}
Here, $\phi_{\ell}$ denotes the corresponding normalization and projection, and $a_{\ell}$ is a fusion coefficient before softmax. These transformations belong to the prepared video-feature interface. Following the frozen-module configuration in Fig.~\ref{fig:DiffWAM_Internal_model_architecture}, the representation front end is held fixed during motion-readout training; no predefined semantic role is assigned to an individual selected depth.
% 中文翻译：每个层级经过独立归一化和从 5,376 维到 512 维的投影，再以 softmax 归一化系数融合。这些变换属于已准备的视频特征接口。依照详细架构图中的冻结配置，表征前端在运动读出训练阶段保持固定，不预先为某个所选层级指定特定语义作用。

Four factorized transformer blocks then apply spatial and temporal self-attention:
\begin{equation}
F=\mathcal{B}_{\bar{\xi}}(F^{\mathrm{fuse}}),
\label{eq:diffwam_factorized_features}
\end{equation}
where $\bar{\xi}$ denotes the fixed parameters of the factorized spatiotemporal module. All associations below use this processed grid $F$. Spatial pooling is performed only after Grid-Motion, so that the association operators can access the unpooled spatial layout.
% 中文翻译：随后，四个分解式 Transformer 模块执行空间和时间自注意力，其参数保持固定。下文所有关联均使用处理后的网格 F，而不是融合前特征。空间池化只在 Grid-Motion 之后执行，使关联算子能够访问未池化的空间布局。

\paragraph{Grid-Motion association.}
For a predictive grid at video-slot index $u$, temporal association retrieves features across spatial locations from reference grids at $\max(0,u-1)$ and $\max(0,u-4)$. The retrieved features are indexed on the current grid, giving
\begin{equation}
M_u^{\Delta}
=
\Psi\!\left(
F_u-\mathcal{A}^{\Delta}_{\theta}
\big(F_u,F_{\max(0,u-1)},F_{\max(0,u-4)}\big)
\right).
\label{eq:grid_motion_temporal}
\end{equation}
Here, $\mathcal{A}^{\Delta}_{\theta}$ performs cross-location retrieval and aggregation, and $\Psi$ transforms the resulting feature difference. The index $u$ refers to a predictive video slot, whereas $t$ indexes an output trajectory pose. The temporal offsets above are measured in feature slots, not seconds or denoising evaluations.
% 中文翻译：对于预测视频槽索引 u 对应的网格，时序关联从 max(0,u-1) 和 max(0,u-4) 对应的参考网格跨空间位置检索特征，并将结果对齐到当前网格。当前特征与检索特征之差经变换后形成时序消息。u 表示预测视频槽，t 表示输出轨迹位姿；这里的时间偏移以特征槽计量，不是秒数，也不是去噪前向次数。

A second association links predictive features to the observed scene. Queries are derived from predictive features and keys from clean observation-anchor features, while values incorporate first-frame geometry. Geometry therefore conditions the aggregated content rather than directly entering the query--key matching scores in this association. We denote this interface by
% 中文翻译：第二组关联将预测特征连接到观测场景。Query 来自预测特征，Key 来自干净观测锚点特征，Value 则融合首帧几何。因此，在这一关联操作中，几何用于条件化被聚合的内容，而不直接进入 Query--Key 匹配分数的计算。该接口表示为
\begin{equation}
K_0=\kappa(A_0),
\qquad
V_0=\nu(A_0,G_0),
\qquad
M_u^{A}=\mathcal{A}^{A}_{\theta}(F_u,K_0,V_0).
\label{eq:diffwam_anchor_association}
\end{equation}
The geometry-conditioned value mapping $\nu$ associates geometric information with the anchor indexing; the geometry grid and the video backbone feature grid need not have identical native resolutions. The two messages are combined with the processed predictive feature through a gated residual:
\begin{equation}
\widetilde{F}_u
=
F_u+\gamma_u^{\Delta}M_u^{\Delta}+\gamma_u^{A}M_u^{A}.
\label{eq:grid_motion_fusion}
\end{equation}
where $\gamma_u^{\Delta}$ and $\gamma_u^{A}$ are learned gates.
% 中文翻译：第二组关联将预测特征连接到观测场景，其 key 来自干净锚点特征，value 融合首帧几何。几何条件化映射将几何信息对应到锚点索引，因此不要求几何网格和 video backbone 特征网格具有相同原生分辨率。两个关联消息与处理后的预测特征通过可学习门控残差融合。
% AUTHOR CHECK 4：上述关联公式保留功能接口，不擅自指定实现细节。
% 请补充两个时间参考的实际组合方式、空间/时间位置编码、kappa 与 nu 的具体实现及网格对齐方式，以及门控的形状与生成方式。

Learned memory pooling maps each $\widetilde{F}_u$ to eight pooling outputs, corresponding to the eight per-slot pooling channels shown in Fig.~\ref{fig:DiffWAM_Internal_model_architecture}. Each output forms one motion-memory token, giving
\begin{equation}
\mathcal{M}=\operatorname{Pool}_{\theta}(\widetilde{F})
\in\mathbb{R}^{312\times512},
\qquad 312=39\times8.
\label{eq:diffwam_motion_memory}
\end{equation}
The eight pooling outputs are distinct from the 512-dimensional feature channels. The separately retained observation anchors are used for association and are not included in this count of pooled predictive tokens.
% 中文翻译：学习式记忆池化将每个时间位置映射为八个池化输出，对应图中每槽八个池化通道；每个输出构成一个运动记忆 token，因此 39 个时间槽共得到 312 个 token。八个池化输出不同于 512 维的特征通道。单独保留的观测锚点用于关联，不计入这 312 个池化预测 token。

These associations are learned soft correspondences rather than measured optical flow. The readout does not explicitly impose orbit templates, target-center constraints, manually specified trajectory shapes, or forced loop closure. The architecture therefore provides learnable motion associations without explicitly imposing task-specific geometric templates.
% 中文翻译：这些关联是学习得到的软对应，而不是实测光流。读出器不显式施加环绕模板、目标中心约束、人工轨迹形状或强制闭环。因此，该架构提供可学习的运动关联，而不显式施加任务专用几何模板。

% =============================================================================
\subsection{Geometry-Guided Trajectory Decoding}
% 中文翻译：几何引导的轨迹解码
\label{sec:architecture}

Predictive motion features do not by themselves define an explicit metric spatial reference. We therefore condition the readout on geometry estimated from $I_0$ by frozen MoGe2. The cached $8\times14$ geometry grid contains signed-log XYZ coordinates, surface normals, log depth, and a validity indicator, giving eight channels per cell. After spatial resampling when required, the geometry MLP embeds these channels. For the coordinate embedding, the signed-log coordinates are decoded to XYZ and normalized by the scene-level scale $s_0$ defined below:
\begin{equation}
\bar{\mathbf{x}}_j
=
\operatorname{clip}\!\left(
\frac{\mathbf{x}_j}{s_0},-20,20
\right).
\end{equation}
The Fourier embedding concatenates $\bar{\mathbf{x}}_j$ with $\sin(f\bar{\mathbf{x}}_j)$ and $\cos(f\bar{\mathbf{x}}_j)$ for $f\in\{1,2,4\}$, yielding 21 coordinate features. The projected coordinate embedding is added to the geometry-MLP output. This normalization uses a shared scene scale rather than each cell's individual depth.
% 中文翻译：预测运动特征本身不能定义显式米制空间参考，因此读出器以冻结的 MoGe2 从初始图像估计的几何为条件。缓存的 8×14 几何网格包含带符号对数 XYZ、表面法向、对数深度及有效性标记，每个单元共八通道。必要的空间重采样完成后，几何 MLP 对这些通道进行嵌入。坐标嵌入先将带符号对数坐标还原为 XYZ，再除以下文定义的场景级尺度 s_0，并逐分量裁剪至 [-20,20]。Fourier 嵌入拼接归一化坐标及频率为 1、2、4 的正弦和余弦特征，共得到 21 维坐标特征，其投影结果与几何 MLP 输出相加。这里使用全局共享的场景尺度归一化，而不是分别除以各单元自身深度。

Let $g_j^{\mathrm{depth}}=\log d_j$ denote the log depth of cell $j$. From the valid cells $\mathcal{V}$, we compute
\begin{equation}
s_0
=
\operatorname{clip}\!\left(
\operatorname{median}_{j\in\mathcal{V}}
\exp(g_j^{\mathrm{depth}}),\,0.1,\,100
\right).
\label{eq:scene_scale}
\end{equation}
The bounds are expressed in meters. When no valid depth estimate is available, the implementation uses unit scale. Thus, $s_0$ provides a metric-scale condition derived either from monocular geometry or from an external depth sensor, depending on the available depth source.
% 中文翻译：上述边界以米为单位。当不存在有效深度估计时，实现中使用单位尺度。因此，$s_0$ 提供米制尺度条件，其既可以由单目几何估计得到，也可以由外部深度传感器提供，具体取决于可用的深度来源。

The decoder contains 38 learned future-pose queries and eight transformer decoder blocks. Each block applies query self-attention, cross-attention to $\mathcal{M}$, cross-attention to the first-frame geometry tokens, and a feed-forward sublayer. Let $q_t$ be the final query representation for timestamp $\delta_t$. The position and rotation heads output
\begin{equation}
\hat{p}_t=s_0(W_pq_t+b_p),
\qquad
\hat{R}_t=\operatorname{GS}(W_Rq_t+b_R),
\label{eq:metric_head}
\end{equation}
where the heads produce three translation coordinates and a six-dimensional rotation representation, respectively. Gram--Schmidt orthogonalization, denoted by $\operatorname{GS}$, converts the latter into a rotation matrix.
% 中文翻译：解码器包含 38 个可学习未来位姿 query 和 8 个 Transformer 解码模块。每个模块依次执行 query 自注意力、运动记忆交叉注意力、首帧几何交叉注意力和前馈计算。位置预测头输出三维平移并乘以场景尺度；旋转预测头输出六维表征，再通过 Gram--Schmidt 正交化得到旋转矩阵。

The video model provides predictive representations of future scene evolution and camera motion, while monocular geometry estimation or an external depth sensor anchors these representations to the geometry and metric scale of the current observation. Their combination enables DiffWAM to predict camera trajectories in the initial camera coordinate frame. The scale factor $s_0$ explicitly conditions translation on scene scale, but does not impose strict scale equivariance because the learned representations themselves also depend on geometric inputs.

% 中文翻译：视频模型提供关于未来场景演化和相机运动的预测表征，而单目几何估计或外部深度传感器则将这些表征锚定到当前观测的几何结构和米制尺度上。二者结合，使 DiffWAM 能够在初始相机坐标系下预测相机轨迹。尺度因子 $s_0$ 显式地使平移预测以场景尺度为条件，但由于学习得到的表征本身也依赖几何输入，因此并不构成严格的尺度等变约束。

\FloatBarrier

% =============================================================================
\subsection{Geometry-Guided Predictive Distillation}
% 中文翻译：几何引导的预测蒸馏
\label{sec:distillation}

\paragraph{Training stages and parameter scope.}
Figure~\ref{fig:DiffWAM_Pipeline} distinguishes video-backbone preparation (S0), navigation-aware readout pretraining (S1), and geometry-guided predictive distillation (GPD, S2). These stage labels are distinct from denoising schedule indices. S1 produces the navigation-aware readout initialization $\theta_{\mathrm{pre}}$. Starting from this initialization, S2 fine-tunes the DiffWAM and DiffWAM-Flash readouts using their respective retained predictive-state subsets. During S1 and S2, the video backbone and the MoGe2 geometry estimator remain frozen, while the trainable motion-readout components are optimized jointly. The geometry embeddings inside the readout are trainable components and are not part of the frozen MoGe2 estimator.
% 中文翻译：总体流程图区分视频骨干准备（S0）、导航感知读出预训练（S1）以及几何引导的预测蒸馏 GPD（S2）。这些训练阶段标签与去噪调度状态编号彼此独立。S1 首先得到导航感知的读出初始化参数 $\theta_{\mathrm{pre}}$；随后，S2 从该初始化出发，分别利用 DiffWAM 和 DiffWAM-Flash 所对应的预测状态子集对读出器进行进一步训练。在 S1 和 S2 中，视频骨干与 MoGe2 几何估计器始终保持冻结，而运动读出器中的可训练组件联合优化。读出器内部的几何嵌入属于可训练组件，并不属于冻结的 MoGe2 几何估计器。
\paragraph{Navigation-aware readout pretraining (S1).}
We initialize the readout using trajectory supervision from the same corpus employed for video-backbone adaptation, without introducing additional scene or instruction data. The extracted predictive features and first-frame geometry are mapped to the available corpus trajectories, producing an initialization $\theta_{\mathrm{pre}}$. This stage establishes a corpus-supervised motion readout. The following stage adds an explicit correspondence between each sampled early predictive state and the reconstructed motion of its own completed latent rollout.
% 中文翻译：导航感知读出预训练 S1。我们使用视频骨干适配所用同一数据集中的轨迹监督初始化读出器，不引入额外场景或指令数据。预测特征和首帧几何被映射至数据集中的轨迹，得到预训练参数。下一阶段进一步建立每个采样早期状态与其自身完整 latent rollout 重建轨迹之间的显式对应。

\paragraph{Geometry-guided predictive distillation (S2).}
GPD transfers trajectory supervision from completed video rollouts to a motion readout operating on early predictive features. Each training example pairs features with a reference trajectory reconstructed from the same rollout. The frozen video model generates these examples offline, independently of the current readout parameters.
% 中文翻译：GPD 将完整视频 rollout 所提供的轨迹监督迁移到使用早期预测特征的运动读出器。每个训练样本中的特征与参考轨迹来自同一次 rollout。冻结视频模型离线生成这些样本，其生成过程不依赖当前读出器参数。

For $(I_0,c)$ and sampled video noise $\epsilon_v$, we retain the feature collection required by the deployed variant:
\begin{equation}
(Z_{\epsilon_v},A_0)
=\mathcal{E}_{\psi}^{\mathcal{K}}(I_0,c,\epsilon_v),
\label{eq:opd_predictive_state}
\end{equation}
where $\mathcal{K}=\{0,1,2,3\}$ for DiffWAM and $\mathcal{K}=\{0\}$ for DiffWAM-Flash. During offline data construction, we complete the same video rollout, including the remaining backbone computation, scheduler updates, and video decoding, to obtain
\begin{equation}
V_{\epsilon_v}
=\operatorname{Rollout}_{\psi}(I_0,c,\epsilon_v).
\label{eq:opd_rollout_video}
\end{equation}
The rollout operator denotes continuation of the same sampled generation rather than an independent video sample. Any additional randomness introduced during continuation is retained to preserve the correspondence between predictive features and the completed video.
% 中文翻译：给定初始图像、指令和视频噪声，我们保存部署版本所需的特征集合：DiffWAM 使用调度状态 {0,1,2,3}，DiffWAM-Flash 使用 {0}。离线数据构建继续完成同一次视频 rollout，包括剩余骨干计算、调度器更新和视频解码，得到完整视频。该过程延续同一次采样生成，而不是另外独立生成视频；后续生成若引入额外随机性，也应保留相同随机实现，以保证预测特征与完整视频对应。

Each record preserves
\begin{equation}
\big(Z_{\epsilon_v},A_0,G_0,V_{\epsilon_v}\big),
\label{eq:opd_paired_record}
\end{equation}
with $G_0=\mathcal{G}_{\eta}(I_0)$ computed from the initial RGB image, matching deployment. Geometry extracted from generated future images, when used by the teacher, is not part of the student's input. Different noise samples can produce different plausible motions for the same $(I_0,c)$; pairing features and trajectories from unrelated rollouts can therefore introduce inconsistent supervision. The preserved correspondence is
\begin{equation}
(Z_{\epsilon_v},A_0)\;\longleftrightarrow\;V_{\epsilon_v},
\label{eq:same_rollout_correspondence}
\end{equation}
so that the decoder is supervised against the motion associated with its own feature realization.
% 中文翻译：每条记录保存早期特征、观测锚点、首帧几何和同源完整视频。学生的首帧几何始终由初始 RGB 图像计算，与部署一致；教师可能使用的未来图像几何不属于学生输入。相同观测和指令在不同噪声下可能产生不同合理运动，因此不能将不相关 rollout 的特征和轨迹任意配对。

\paragraph{Reconstruction-derived, scale-calibrated teacher.}
$\pi^3$ reconstructs camera motion and depth from the completed video. Following the MoGe2-calibrated teacher in Fig.~\ref{fig:DiffWAM_Internal_model_architecture}, we estimate its translation scale using MoGe2 depth for the corresponding calibration images. Let $\mathcal{J}_{\mathrm{cal}}$ denote the selected calibration-frame indices. For $t\in\mathcal{J}_{\mathrm{cal}}$, $D^{\mathrm{metric}}_{t,j}$ is the MoGe2-estimated depth and $D^{\mathrm{\pi^3}}_{t,j}$ is the $\pi^3$ depth at the corresponding image pixel. These depths must refer to the same image and registered pixel locations; unrelated future sensor observations cannot substitute for generated-image depth.
% 中文翻译：重建与尺度校准教师。$\pi^3$ 从完整视频恢复相机运动和深度。根据图中的 MoGe2 定标教师，我们利用对应定标图像的 MoGe2 深度估计平移尺度。定标帧集合中的两种深度必须来自同一图像及已配准像素位置，不能使用不相关未来传感器观测的深度替代生成图像深度。

For finite depth pairs satisfying
\begin{equation}
0.5<D^{\mathrm{metric}}_{t,j}<30,
\qquad D^{\mathrm{\pi^3}}_{t,j}>0,
\label{eq:diffwam_teacher_depth_validity}
\end{equation}
we form the valid set $\mathcal{V}_{\mathrm{teacher}}$ and estimate
\begin{equation}
\alpha
=\operatorname{median}_{(t,j)\in\mathcal{V}_{\mathrm{teacher}}}
\frac{D^{\mathrm{metric}}_{t,j}}{D^{\mathrm{\pi^3}}_{t,j}}.
\label{eq:teacher_scale}
\end{equation}
The metric-depth bounds are in meters. This definition requires a successful reconstruction, a nonempty valid set, and a finite positive $\alpha$. If calibration uses only the initial image, then $\mathcal{J}_{\mathrm{cal}}=\{0\}$; using additional generated frames changes the teacher's information budget but not the student's first-frame-only input.
% 中文翻译：对于满足原稿深度范围及正值条件的有限深度对，取有效深度比值的中位数作为统一尺度因子。该定义要求重建成功、有效集合非空且尺度为有限正值。若仅使用初始图像定标，则定标帧集合为 {0}；使用额外生成帧只改变教师的信息范围，不改变学生仅访问首帧几何的约束。
% AUTHOR CHECK 7：图中确定教师由 MoGe2 定标，但没有确定定标使用首帧、部分帧还是所有帧。
% 请明确 J_cal、像素配准/有效性规则，以及重建或尺度估计失败时的实际过滤或重生成策略；这里不虚构具体实现。

Writing the reconstructed poses in camera-to-reconstruction-frame convention as $(\bar{R}_t,\bar{p}_t)$, we express the supervision in the initial camera frame:
\begin{equation}
R_t^{\star}=\bar{R}_0^{\top}\bar{R}_t,
\qquad
p_t^{\star}=\alpha\bar{R}_0^{\top}(\bar{p}_t-\bar{p}_0).
\label{eq:diffwam_teacher_alignment}
\end{equation}
Thus, $(R_0^{\star},p_0^{\star})=(I_3,\mathbf{0})$, and the same $\alpha$ scales all translation axes. Scale calibration does not change rotations; the rotation above only changes the reference frame. If reconstruction poses are provided in the inverse convention, they are first converted to the convention used in this equation. Teacher and student poses correspond to the same time offsets $\delta_t$.
% 中文翻译：将重建位姿写为相机到重建参考系的变换后，利用初始位姿把所有监督转换到初始相机坐标系，并使用同一个尺度因子缩放全部平移轴。尺度定标不改变旋转；式中的旋转变换仅用于改变参考坐标系。若重建器提供相反方向的外参，需先转换约定。教师和学生使用相同时间偏移。

The resulting trajectory is
\begin{equation}
\tau_{\epsilon_v}^{\star}
=\mathcal{T}(V_{\epsilon_v},D^{\mathrm{metric}}),
\label{eq:teacher_trajectory}
\end{equation}
where $\mathcal{T}$ includes reconstruction, scale calibration, and coordinate and temporal alignment. This is a reconstruction-derived reference in estimated meters, not measured physical ground truth. Same-rollout pairing ensures source correspondence, but does not remove generation errors, reconstruction drift, or monocular-scale bias.
% 中文翻译：完整教师过程包含重建、尺度校准、坐标对齐和时间对齐。所得轨迹是以估计米制尺度表示的重建参考，而不是实测物理真值。同源 rollout 配对保证监督来源对应，但不会消除生成错误、重建漂移或单目尺度偏差。

\paragraph{Distillation objective.}
Starting from $\theta_{\mathrm{pre}}$, the readout predicts
\begin{equation}
\hat{\tau}_{\epsilon_v}
=\mathcal{D}_{\theta}(Z_{\epsilon_v},A_0,G_0),
\qquad \theta\leftarrow\theta_{\mathrm{pre}},
\label{eq:opd_prediction}
\end{equation}
and is optimized through
\begin{equation}
\theta^{\star}
=\arg\min_{\theta}\;
\mathbb{E}_{(I_0,c)\sim\mathcal{D}_{\mathrm{train}},\,\epsilon_v}
\!\left[
\mathcal{L}\!\left(
\mathcal{D}_{\theta}(Z_{\epsilon_v},A_0,G_0),
\tau_{\epsilon_v}^{\star}
\right)
\right].
\label{eq:opd_objective}
\end{equation}
The teacher accesses the completed generated future, whereas the student uses only early predictive features and first-frame geometry. The targets train the readout to recover the teacher-associated motion; they do not update the frozen video backbone representations. This is trajectory-level supervised distillation of an expensive continuation-and-reconstruction process into a direct motion readout.
% 中文翻译：蒸馏目标。读出器以预训练参数初始化，最小化其预测与同源教师轨迹之间的误差。教师访问完整生成未来，学生仅使用早期预测特征和首帧几何。目标用于训练读出器恢复教师对应运动，而不更新冻结的 H3 表征。这是在轨迹层面把昂贵的完整生成与重建过程蒸馏为直接运动读出。

For the $T=38$ future poses, the two-term objective shown in Fig.~\ref{fig:DiffWAM_Internal_model_architecture} is
\begin{equation}
\mathcal{L}=\mathcal{L}_{\mathrm{pos}}+\lambda_R\mathcal{L}_{\mathrm{rot}},
\label{eq:training_objective}
\end{equation}
where
\begin{equation}
\mathcal{L}_{\mathrm{pos}}
=\frac{1}{3T}\sum_{t=1}^{T}\sum_{j=1}^{3}
\rho\!\left(
\frac{\hat{p}_{t,j}-p_{t,j}^{\star}}{\operatorname{sg}(s_0)}
\right),
\label{eq:position_objective}
\end{equation}
and
\begin{equation}
\mathcal{L}_{\mathrm{rot}}
=\frac{1}{T}\sum_{t=1}^{T}
 d_{\mathrm{SO}(3)}(\hat{R}_t,R_t^{\star})^2.
\label{eq:rotation_objective}
\end{equation}
Here, $\rho$ is the robust scalar position penalty and $\operatorname{sg}$ denotes stop-gradient. The rotation discrepancy is the geodesic angle in radians,
\begin{equation}
 d_{\mathrm{SO}(3)}(R_1,R_2)
 =\arccos\!\left[
 \operatorname{clip}\!\left(
 \frac{\operatorname{tr}(R_1^{\top}R_2)-1}{2},-1,1
 \right)\right].
\label{eq:diffwam_rotation_geodesic}
\end{equation}
The fixed initial pose is excluded from supervision, and all future timestamps receive equal weight. The teacher factor $\alpha$ calibrates reconstruction scale, whereas $s_0$ conditions and normalizes the student's translation readout; they have different roles.
% 中文翻译：图示两项损失分别为尺度归一化的鲁棒位置损失和旋转测地角平方损失，监督 38 个未来位姿。旋转测地角使用弧度。固定初始位姿不参与监督，所有未来时间位置具有相同权重。教师尺度因子 alpha 用于重建定标，学生场景尺度 s_0 用于平移条件化与损失归一化，二者作用不同。

The readout and its objectives are shared across instruction categories; task conditions enter through the predictive features rather than through hand-designed trajectory templates. Historical composite-loss configurations discussed in Sec.~\ref{sec:training_results} are separate training variants, not additional unlisted terms in Eq.~\eqref{eq:training_objective}. At deployment, the complete video rollout, $\pi^3$ reconstruction, and teacher-side calibration are removed.
% 中文翻译：读出器及其目标在不同指令类别之间共享；任务条件通过预测特征进入模型，而不是通过手工轨迹模板实现。实验部分的历史辅助监督及复合损失属于独立训练变体，不能视为当前两项目标中未列出的附加项。在部署时，完整视频 rollout、$\pi^3$ 重建和教师侧尺度校准均被移除。

% =============================================================================
\subsection{Inference}
% 中文翻译：推理
\label{sec:method_inference}

At deployment, the external inputs are the initial RGB observation $I_0$ and language instruction $c$. The video backbone and MoGe2 can be evaluated concurrently: the predictive branch depends on $(I_0,c)$ and internally sampled video noise, while the geometry branch depends only on $I_0$. The predictive branch performs only the backbone evaluations and scheduler updates required to reach the final retained schedule state, and terminates immediately after DiT block 35 of the final required evaluation. No further scheduler updates or VAE video decoding are performed. Making the internal noise dependence explicit, the inference path is
% 中文翻译：部署时，外部输入仅包括初始 RGB 观测 $I_0$ 和语言指令 $c$。视频骨干与 MoGe2 可以并行计算：预测分支依赖 $(I_0,c)$ 以及内部采样的视频噪声，而几何分支仅依赖 $I_0$。预测分支只执行到达最后一个保留调度状态所必需的骨干前向与调度器更新，并在最后一次所需前向的 DiT 第 35 号块后立即终止。此后不再执行额外调度器更新，也不进行 VAE 视频解码。显式写出内部噪声依赖后，推理路径为
\begin{equation}
(I_0,c;\epsilon_v)
\longrightarrow
(Z_{\epsilon_v},A_0,G_0)
\longrightarrow
\hat{\tau}_{\epsilon_v}.
\label{eq:inference_path}
\end{equation}
No completed future video, online $\pi^3$ reconstruction, or teacher trajectory is required.
% 中文翻译：部署时，外部输入仅为初始 RGB 观测和语言指令。预测分支依赖观测、指令及内部视频噪声，几何分支仅依赖同一初始图像，因此二者可以并行。H3 经提前退出接口提供所选特征，不执行后续调度更新或 VAE 视频解码。在线路径不需要完整未来视频、$\pi^3$ 重建或教师轨迹。

The predicted pose sequence is a motion proposal, not a certified collision-free vehicle command. Its initial-camera reference and pose time offsets are preserved when calibrated camera-to-body extrinsics and the downstream state estimate map it to the execution frame. The planner checks feasibility and constructs an executable trajectory, while the control stack handles tracking and vehicle constraints. This separation lets DiffWAM focus on language-conditioned geometric motion prediction without attributing collision avoidance or dynamic-feasibility guarantees to the pose decoder itself.
% 中文翻译：预测位姿序列属于运动提议，而不是已认证的无碰撞飞行指令。通过标定的相机到机体外参和状态估计映射到执行参考系时，保留初始相机参考与各位姿的时间偏移。规划器执行可行性检查并构建可执行轨迹，控制系统负责跟踪及飞行器约束。该划分使 DiffWAM 专注于语言条件下的几何运动预测，而不把避障或动力学可行性保证归因于位姿解码器本身。

\section[FastDreamer: Inference and Continuous Execution]{FastDreamer: Inference and Continuous Execution}
% 中文翻译：FastDreamer：端侧推理与连续执行
\label{sec:fastdreamer}

FastDreamer is an inference-and-execution framework designed to connect DiffWAM to continuous UAV control, with particular emphasis on onboard deployment of the DiffWAM-Flash configuration on the Thor platform. It addresses two challenges that become critical in closed-loop execution. First, short-horizon motion prediction benefits from a locally grounded instruction, but maintaining an additional large vision-language model solely for instruction rewriting increases the onboard memory requirement. FastDreamer therefore defines a shared-weight rewriting interface that reuses the vision-language component of the world-model conditioning path. Second, proposal preparation takes time while the UAV continues to move. FastDreamer overlaps this computation with the execution of a currently validated trajectory and aligns each accepted update with a scheduled future handoff state.

% 中文翻译：FastDreamer 是将 DiffWAM 接入连续无人机控制的推理与执行框架，并重点面向 Thor 平台上 DiffWAM-Flash 配置的端侧部署。它主要处理闭环执行中的两个关键问题。第一，短时域运动预测需要与当前场景和局部目标相匹配的指令，但仅为指令改写而额外常驻一个大型视觉语言模型会增加端侧显存需求。因此，FastDreamer 定义了共享权重的改写接口，复用世界模型条件分支中的视觉语言模块。第二，无人机在轨迹准备过程中仍然持续运动。FastDreamer 将下一条轨迹的准备与当前已通过校验的轨迹执行相重叠，并将通过检查的新轨迹与预定未来交接时刻的状态对齐。

Shared-weight rewriting and parallel proposal inference address memory residency and computation dependencies; Flight-Time Compute determines whether a new plan can be prepared within the remaining execution horizon; and Prospective Handoff addresses the state mismatch accumulated during preparation. These mechanisms have distinct evaluation targets. The inference-performance protocol in Sec.~\ref{sec:efficiency} measures model-pipeline latency and memory, while waiting during execution and handoff continuity characterize separate closed-loop properties. Neither low prediction latency nor a feasible scheduling budget alone establishes closed-loop task success.

% 中文翻译：共享权重改写与并行轨迹推理处理模型常驻开销和计算依赖；Flight-Time Compute 判断新计划能否在当前剩余执行时域内准备完成；Prospective Handoff 则处理准备期间累积的状态失配。这些机制对应不同的评估目标。Sec.~\ref{sec:efficiency} 中的推理性能协议测量模型流水线延迟和显存，而执行过程中的等待以及轨迹交接连续性属于不同的闭环系统属性。较低的预测延迟或满足时间预算，本身均不能证明闭环任务成功。

% TODO(author): 本次仅修改 LaTeX 正文与下方 TikZ 时序图；
% 外部架构 PDF 需同步更新：
% 1. 改写输入加入任务进度 h_n，算子改为 Q_omega；
% 2. MoGe2 与 prompt rewriting 同时启动，而不是等待改写完成；
% 3. 将 Camera -> world -> body 改为世界系相机位姿到世界系机体位姿；
% 4. 删除未定义的 confidence，统一为时间戳、任务/计划标识及校验状态；
% 5. 更新 L_proposal、预定 t_{h,n}、回退余量 R_n 和进度感知连接说明；
% 6. 当前模型接口为 36 个未来位姿加初始位姿，共 37 个位姿。

\subsection{Shared-Weight Prompt Rewriting and Parallel Proposal Inference}
% 中文翻译：共享权重的提示词改写与并行轨迹提议推理
\label{sec:direct}

\paragraph{Shared-weight prompt rewriting.}
A mission-level instruction may contain multiple stages or completion requirements that cannot be inferred from a single current image. For long-horizon execution, the rewriting interface therefore requires task context $h_n$ from an upstream mission manager, in addition to the mission instruction $\mathcal I$ and observation $I_n$. This context describes the active subtask and verified execution progress, such as completed stages or remaining repetitions. At replanning cycle $n$, the local condition is
\begin{equation}
c_n = \mathcal Q_{\omega}(I_n,\mathcal I,h_n),
\label{eq:prompt_rewrite}
\end{equation}
where $\mathcal Q_{\omega}$ denotes the rewriter and $\omega$ denotes the reused vision-language parameters. This notation distinguishes rewriting from the complete video backbone rollout operator used for offline supervision. For an independent single-stage request, $h_n$ can be empty. Updating task progress and declaring mission completion remain responsibilities of the mission manager, not of frame-wise rewriting alone.

% 中文翻译：共享权重的提示词改写。任务级指令可能包含多个阶段，或者无法仅从当前图像推断的完成要求。因此，长时域执行中的改写接口除任务指令和当前观测外，还需要上层任务管理器提供任务上下文 h_n，包括当前子任务以及已经确认的执行进度，例如已完成阶段或剩余重复次数。局部条件由 c_n=Q_omega(I_n,I,h_n) 给出，其中 Q_omega 表示提示词改写器，omega 表示复用的视觉语言参数；该符号与离线监督中的完整 H3 rollout 算子区分。独立单阶段请求的 h_n 可以为空。任务进度更新和完成判定由任务管理器负责，而不由逐帧提示词改写独立完成。

The rewriter is instructed to preserve the target identity, commanded side, distance, altitude, motion direction, and temporal constraints relevant to the active subtask. Global stage ordering and completion requirements remain in the task context rather than being reset at each local prediction. Prompt rewriting~\cite{rewrite} adapts the user instruction to the conditioning distribution of the video model by making the intended motion and scene relations more explicit. Prior work has shown that such prompt optimization can improve instruction alignment in generated videos. In FastDreamer, the rewritten condition is used directly by the predictive backbone, providing a more explicit motion condition for the downstream trajectory readout without requiring future-video generation.

% 中文翻译：改写器被要求保留与当前子任务相关的目标身份、指定侧向、距离、高度、运动方向和时间约束。全局阶段顺序和完成要求由任务上下文持续维护，而不是在每次局部预测时重置。提示词改写~\cite{rewrite} 通过显式化预期运动和场景关系，使用户指令更适配视频模型的条件输入分布。已有工作表明，这类提示词优化能够提升生成视频与输入指令之间的对齐程度。在 FastDreamer 中，改写后的条件直接输入预测骨干网络，为下游轨迹读出提供更加明确的运动条件，而无需实际生成未来视频。

\paragraph{Parallel proposal inference.}
Once the request inputs are available, first-image geometry estimation can start concurrently with prompt rewriting because MoGe2 does not depend on $c_n$. After rewriting, the prediction branch computes image-language conditioning from $(I_n,c_n)$ and executes the predictive-backbone evaluations required by the deployed DiffWAM variant, terminating after the deepest retained feature layer of the final required evaluation. DiffWAM-Flash uses a single truncated evaluation, whereas the standard DiffWAM configuration retains multiple predictive evaluations as defined in Sec.~\ref{sec:method}. The trajectory decoder runs after both the predictive features and geometry features are available. Both branches use the same captured image, and neither configuration performs future-video decoding or online $\pi^3$ reconstruction.

% 中文翻译：并行轨迹提议推理。请求输入准备好后，由于 MoGe2 不依赖局部条件 $c_n$，首帧几何估计可以与提示词改写同时启动。完成改写后，预测分支根据 $(I_n,c_n)$ 计算图像语言条件，并执行当前 DiffWAM 配置所需的预测骨干网络计算，在最后一次所需预测计算的最深保留特征层处终止。DiffWAM-Flash 仅执行一次截断的预测计算，而标准 DiffWAM 配置按照 Sec.~\ref{sec:method} 中的定义保留多次预测计算。轨迹解码器在预测特征与几何特征均就绪后运行。两条分支使用同一张拍摄图像，两种配置均不执行未来视频解码或在线 $\pi^3$ 重建。

Let $T_{\rm pred}$ denote the duration of the prediction branch, including rewriting, and $T_{\rm geo}$ the duration of the geometry branch:
\begin{equation}
T_{\rm pred}
=
T_{\rm rewrite}
+
T_{\rm video\ condition}
+
T_{\rm video\ DiT}
+
T_{\rm feature\ transfer},
\label{eq:predictive_branch_latency}
\end{equation}
\begin{equation}
T_{\rm geo}
=
T_{\rm MoGe2}
+
T_{\rm geometry\ transfer}.
\label{eq:geometry_branch_latency}
\end{equation}
The idealized model-proposal inference critical path is
\begin{equation}
T_{\rm proposal}^{\rm ideal}
=
T_{\rm input}
+
\max(T_{\rm pred},T_{\rm geo})
+
T_{\rm head}
+
T_{\rm output},
\label{eq:criticalpath}
\end{equation}
where $T_{\rm input}$ covers request-side input handling and $T_{\rm output}$ covers delivery of the predicted proposal to the planning interface. This quantity includes rewriting but excludes downstream transition planning and validation. It is therefore one component of the complete preparation latency defined in Sec.~\ref{sec:rewrite}, and is distinct from the model-only latency measured with a prepared image and local condition.

% 中文翻译：预测分支耗时包含提示词改写、视频模型条件计算、视频骨干网络预测计算和预测特征传输；几何分支耗时包含 MoGe2 推理和几何特征传输。理想的模型轨迹提议推理关键路径由请求侧输入处理、两条并行分支中较慢的一条、轨迹解码器和提议交付组成。该量已经包含提示词改写，但不包含下游的过渡规划和校验。因此，它只是 Sec.~\ref{sec:rewrite} 所定义完整准备时延的一部分，同时也不同于从已经准备好的图像和局部条件开始统计的纯模型延迟。

\subsection{Flight-Time Compute}
% 中文翻译：飞行时间计算预算
\label{sec:rewrite}

DiffWAM predicts a finite-horizon motion proposal. FastDreamer uses the remaining duration of the currently validated committed trajectory as a preparation budget, rather than waiting for that trajectory to end before requesting an update. We distinguish five timestamps: image capture $t_n$, request launch $u_n$, preparation completion $r_n$, scheduled handoff $t_{h,n}$, and expiry of the current committed trajectory $e_n$. Preparation includes proposal inference, transition planning, validation, and delivery. For a timely accepted update, these timestamps satisfy
\[
t_n
\leq u_n
\leq r_n
\leq t_{h,n}
\leq e_n.
\]
The measured preparation duration is $d_n=r_n-u_n$, whereas the observation-to-handoff age is $\Delta_n=t_{h,n}-t_n$. These durations need not be equal.

% 中文翻译：DiffWAM 预测有限时域的运动提议。FastDreamer 将当前已通过校验并提交的轨迹的剩余执行时间作为准备预算，而不是等待轨迹结束后才请求更新。这里区分图像拍摄、请求发起、准备完成、预定交接和当前轨迹到期五个时刻。准备过程包含轨迹推理、过渡规划、校验和交付。对于按时接受的更新，时间戳满足 t_n<=u_n<=r_n<=t_{h,n}<=e_n。准备耗时 d_n=r_n-u_n 与观测到交接的年龄 Delta_n=t_{h,n}-t_n 不一定相等。

The available flight-time budget is
\begin{equation}
B_n=e_n-u_n.
\label{eq:flight_time_budget}
\end{equation}
An additive accounting of the preparation stages is
\begin{equation}
\widehat L_n
=
\widehat L_{{\rm proposal},n}
+
\widehat L_{{\rm planner},n}
+
\widehat L_{{\rm validation},n}
+
\widehat L_{{\rm communication},n}^{\rm extra},
\label{eq:estimated_prepare_latency}
\end{equation}
where the proposal estimate includes rewriting and parallel geometry. The last term includes only communication not already accounted for in input handling, feature transfer, or proposal delivery. Validation and activation checks are explicitly budgeted rather than hidden in an unspecified margin. When these stages overlap, the estimate follows the measured critical path instead of summing overlapping intervals or stage-wise latency quantiles.

% 中文翻译：可用飞行时间预算为 B_n=e_n-u_n。准备时延按轨迹提议、规划、校验及额外通信进行核算，其中轨迹提议部分已经包含提示词改写和并行几何。额外通信只统计尚未包含在输入处理、特征传输和提议交付中的部分。校验与激活检查显式计入预算，不再隐藏在未定义的余量中。若这些阶段也存在重叠，则根据实测关键路径估计，而不重复累加重叠区间，也不直接累加各阶段的延迟分位数。

Let $M_n\geq0$ account for uncertainty in the preparation-time estimate, and let $R_n\geq0$ reserve the time required by the flight stack to enter a validated fallback before $e_n$. The scheduling condition is
\begin{equation}
\widehat L_n+M_n+R_n\leq B_n.
\label{eq:budget}
\end{equation}
The handoff time is selected before transition planning, with
\[
u_n+\widehat L_n+M_n
\leq t_{h,n}
\leq e_n-R_n.
\]
State prediction and transition planning use this same scheduled time. An early result waits for the scheduled activation; a result that misses that time must be rescheduled and revalidated or rejected, rather than activated with an outdated boundary condition. Equation~\ref{eq:budget} is an admission condition based on estimated latency, not a deterministic runtime guarantee.

% 中文翻译：M_n 用于覆盖准备耗时估计的不确定性，R_n 则为飞行系统在 e_n 之前进入已经校验的回退行为预留时间。只有预计准备耗时、时间不确定性余量和回退余量之和不超过预算时，才满足调度条件。交接时刻在过渡规划之前选定，状态预测和规划使用同一个预定时刻。提前完成的结果等待预定激活；错过该时刻的结果必须重新调度并校验，或者被拒绝，不能继续使用过时的边界条件直接激活。该条件基于时延估计，并非运行时确定性保证。

The fallback decision cannot be postponed until the committed trajectory has expired. If no usable update is available by $e_n-R_n$, the flight stack must retain a validated continuation or initiate its feasible braking or holding behavior. A zero reserve is admissible only when the committed trajectory already provides the required terminal behavior. Changes that invalidate the committed trajectory may require an earlier response, independently of the nominal compute budget.

% 中文翻译：回退决策不能推迟到当前轨迹到期后。如果在 e_n-R_n 之前仍无可用更新，飞行系统必须保留已经校验的延续轨迹，或者启动可行的制动或保持行为。只有当前轨迹已经包含所需的终端行为时，才允许零回退余量。若环境或状态变化导致当前轨迹失效，则可能需要早于名义预算截止时间作出响应。

For completed preparation attempts, the overrun beyond the committed-plan expiry is
\begin{equation}
g_n=\max(0,d_n-B_n).
\label{eq:uncovered_time}
\end{equation}
The condition $g_n=0$ means only that preparation did not extend beyond $e_n$; it does not imply that the earlier scheduled handoff was met or that the proposal was accepted. Preparation latency is hidden from execution only when a valid, feasible update is ready for the scheduled handoff and ongoing motion remains executable. This reduces waiting caused by computation, not the intrinsic model latency. Rejected proposals, missed handoffs, and unfinished requests must be recorded separately rather than counted as successful zero-wait updates. Requests must also retain their mission and committed-plan identity so that superseded results cannot overwrite newer decisions.

% 中文翻译：对于已经完成的准备尝试，$g_n$ 表示准备过程超出当前已提交轨迹到期时刻的时间。$g_n=0$ 仅说明准备没有晚于 $e_n$，并不意味着满足了更早的预定交接时间，也不意味着提议被接受。只有有效且可行的更新在预定交接时刻就绪，同时当前运动仍然可执行时，准备耗时才不会表现为执行等待。这减少的是计算导致的等待，而不是模型固有延迟。被拒绝的提议、错过交接的提议和未完成请求需分别记录，不能记作成功的零等待更新。请求还必须保留任务和已提交计划的身份信息，以防已经被后续决策替代的旧结果覆盖新的决策。

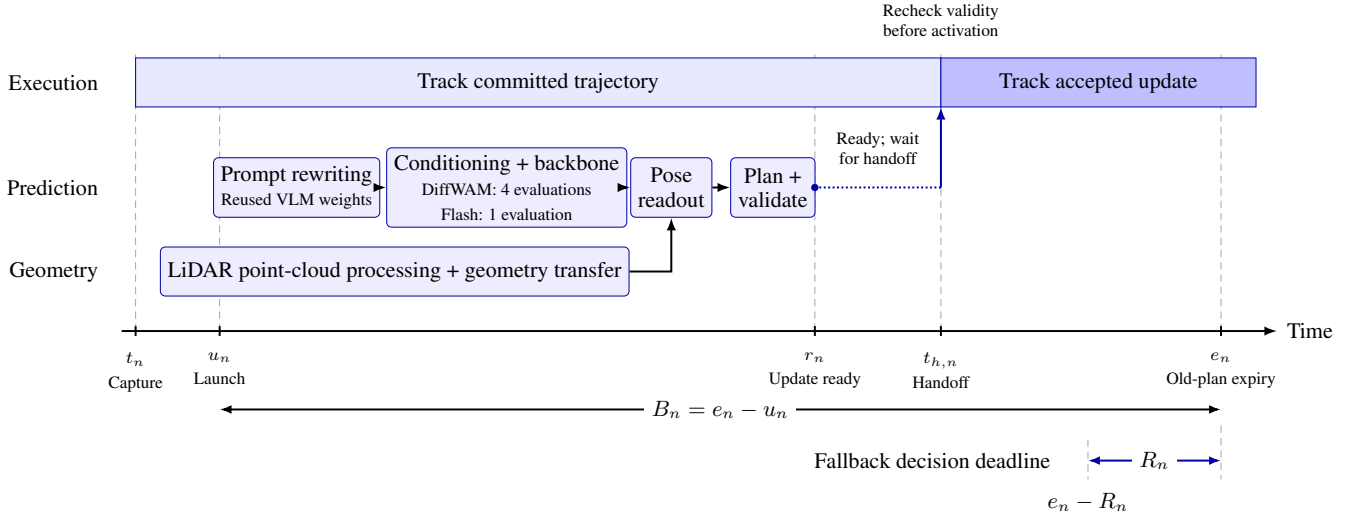
\begin{figure}[htbp]
\centering
\resizebox{\linewidth}{!}{%
\begin{tikzpicture}[
    x=1cm,
    y=1cm,
    font=\small,
    flow/.style={
        -{Latex[length=1.8mm]},
        thick
    },
    timing/.style={
        densely dashed,
        black!35
    },
    module/.style={
        draw=blue!65!black,
        fill=blue!7,
        rounded corners=2pt,
        align=center,
        font=\footnotesize,
        minimum height=0.85cm,
        inner sep=3pt
    },
    event/.style={
        align=center,
        font=\scriptsize,
        anchor=north
    },
    budget/.style={
        {Latex[length=1.7mm]}-{Latex[length=1.7mm]},
        thick
    }
]

% Shared time coordinates.
% 中文翻译：统一使用这些时间坐标，确保事件、标签和预算箭头对齐。
\def\tCapture{0.5}
\def\tLaunch{1.7}
\def\tReady{10.2}
\def\tHandoff{12.0}
\def\tDeadline{14.1}
\def\tExpiry{16.0}

% Time guides are drawn behind all modules.
% 中文翻译：时间辅助线置于模块后方。
\foreach \x in {
    \tCapture,\tLaunch,\tReady,\tHandoff,\tExpiry
}{
    \draw[timing] (\x,0) -- (\x,3.95);
}

% Row labels.
% 中文翻译：流程行标签。
\node[anchor=east] at (0.1,3.55) {Execution};
\node[anchor=east] at (0.1,2.05) {Prediction};
\node[anchor=east] at (0.1,0.85) {Geometry};

% Execution: current trajectory followed by an accepted update.
% 中文翻译：先执行当前轨迹，通过交接检查后切换至新轨迹。
\draw[
    draw=blue!65!black,
    fill=blue!9
]
    (\tCapture,3.2) rectangle (\tHandoff,3.9);

\node at (6.25,3.55)
    {Track committed trajectory};

\draw[
    draw=blue!65!black,
    fill=blue!25
]
    (\tHandoff,3.2) rectangle (16.5,3.9);

\node at (14.25,3.55)
    {Track accepted update};

% Dashed expiry marker refers to the OLD trajectory.
% 中文翻译：到期标记指原有轨迹，而非已经接受的新轨迹。
\node[
    anchor=south,
    align=center,
    font=\scriptsize
] at (\tHandoff,4.08)
    {Recheck validity\\before activation};

% Prediction modules.
% 中文翻译：预测模块。
\node[
    module,
    minimum width=2.2cm
] (rewrite) at (2.8,2.05)
    {Prompt rewriting\\
     {\scriptsize Reused VLM weights}};

\node[
    module,
    minimum width=3.0cm
] (backbone) at (5.8,2.05)
    {Conditioning + backbone\\
     {\scriptsize DiffWAM: 4 evaluations}\\
     {\scriptsize Flash: 1 evaluation}};

\node[
    module,
    minimum width=1.0cm
] (head) at (8.15,2.05)
    {Pose\\readout};

\node[
    module,
    minimum width=1.2cm
] (plan) at (9.6,2.05)
    {Plan +\\validate};

% LiDAR geometry preparation starts in parallel.
% 中文翻译：实机雷达几何处理与提示词改写并行。
\node[
    module,
    minimum width=5.0cm,
    minimum height=0.7cm
] (geometry) at (4.2,0.85)
    {LiDAR point-cloud processing + geometry transfer};

% Dependencies.
% 中文翻译：模块依赖关系。
\draw[flow] (rewrite.east) -- (backbone.west);
\draw[flow] (backbone.east) -- (head.west);
\draw[flow] (geometry.east) -| (head.south);
\draw[flow] (head.east) -- (plan.west);

% The candidate is ready at r_n and waits until handoff.
% 中文翻译：候选轨迹在 r_n 就绪，等待预定交接。
\draw[densely dotted,thick,blue!65!black]
    (\tReady,2.05) -- (\tHandoff,2.05);

\fill[blue!65!black]
    (\tReady,2.05) circle (1.5pt);

\node[
    align=center,
    font=\scriptsize,
    anchor=south
] at (11.1,2.25)
    {Ready; wait\\for handoff};

\draw[flow,blue!65!black]
    (\tHandoff,2.05) -- (\tHandoff,3.2);

% Time axis.
% 中文翻译：时间轴。
\draw[flow]
    (0.3,0) -- (16.8,0)
    node[anchor=west] {Time};

% Event ticks and labels share exactly the same coordinates.
% 中文翻译：刻度、时间符号和事件标签使用相同横坐标。
\foreach \x in {
    \tCapture,\tLaunch,\tReady,\tHandoff,\tExpiry
}{
    \draw[thick] (\x,0.07) -- (\x,-0.07);
}

\node[event] at (\tCapture,-0.17)
    {$t_n$\\[2pt]Capture};

\node[event] at (\tLaunch,-0.17)
    {$u_n$\\[2pt]Launch};

\node[event] at (\tReady,-0.17)
    {$r_n$\\[2pt]Update ready};

\node[event] at (\tHandoff,-0.17)
    {$t_{h,n}$\\[2pt]Handoff};

\node[event] at (\tExpiry,-0.17)
    {$e_n$\\[2pt]Old-plan expiry};

% Remaining execution budget: launch to old-plan expiry.
% 中文翻译：剩余执行预算从请求启动时刻延伸至旧轨迹到期时刻。
\draw[budget]
    (\tLaunch,-1.12) -- (\tExpiry,-1.12);

\node[
    fill=white,
    inner sep=3pt,
    font=\footnotesize
] at (8.85,-1.12)
    {$B_n=e_n-u_n$};

% Fallback reserve: deadline to old-plan expiry.
% 中文翻译：回退预留时间从决策截止点延伸至旧轨迹到期点。
\draw[budget,blue!65!black]
    (\tDeadline,-1.85) -- (\tExpiry,-1.85);

\node[
    fill=white,
    inner sep=3pt,
    font=\footnotesize
] at (15.05,-1.85)
    {$R_n$};

\draw[timing]
    (\tDeadline,-1.55) -- (\tDeadline,-2.12);

\draw[timing]
    (\tExpiry,-1.35) -- (\tExpiry,-2.12);

\node[
    anchor=east,
    font=\footnotesize
] at (13.7,-1.85)
    {Fallback decision deadline};

\node[
    anchor=north,
    font=\footnotesize
] at (\tDeadline,-2.18)
    {$e_n-R_n$};

\end{tikzpicture}%
}

\caption{
\textbf{Asynchronous trajectory preparation and scheduled handoff.}
While the UAV tracks its committed trajectory, prompt rewriting and LiDAR-based geometry preparation proceed in parallel.
The predictive backbone, pose readout, planning, and validation produce a candidate update ready at $r_n$.
DiffWAM uses four backbone evaluations, whereas DiffWAM-Flash uses one; both terminate the final evaluation after DiT block 35 without future-video decoding.
An early candidate waits until the scheduled handoff $t_{h,n}$ and is activated only after its validity is rechecked.
The remaining budget is $B_n=e_n-u_n$, with a fallback decision required no later than $e_n-R_n$ if no valid update can be activated.
The diagram illustrates an early-ready update; horizontal distances are schematic rather than measured durations.
}
% 中文翻译：异步轨迹准备与预定交接。无人机跟踪当前已提交轨迹时，提示词改写与基于雷达的几何准备并行进行。预测骨干、位姿读出、规划和校验产生候选更新，并在 r_n 时刻就绪。DiffWAM 执行四次骨干前向，DiffWAM-Flash 执行一次；两者均在最后一次前向的 DiT 第 35 号块后退出，不解码未来视频。提前就绪的候选轨迹等待至预定交接时刻 t_{h,n}，重新检查有效性后才激活。剩余预算为 B_n=e_n-u_n；若无法激活有效更新，最迟应在 e_n-R_n 时刻作出回退决策。本图示意候选更新提前就绪的情况，水平距离不表示实际测量时间。

\label{fig:async}
\end{figure}
\FloatBarrier

\subsection{Prospective Handoff and Safety Boundary}
% 中文翻译：前瞻式轨迹交接与安全边界
\label{sec:handoff}

\paragraph{Capture-time coordinate grounding.}
A proposal is conditioned on an image captured at $t_n$, not on the vehicle state at its later activation. Let $W$ denote the world frame, $C_n$ the capture-time camera frame, and $k$ a predicted-pose index. Using the estimated capture-time camera pose, the predicted poses are lifted into the world frame as
\begin{equation}
{}^{W}\widehat T_{C_{n,k}}
=
{}^{W}T_{C_n}\,
{}^{C_n}\widehat T_{C_{n,k}}.
\label{eq:world_transform}
\end{equation}
Let ${}^{B}T_C$ be the calibrated rigid transform mapping camera coordinates into body coordinates. The corresponding world-frame body-pose references are
\begin{equation}
{}^{W}\widehat T_{B_{n,k}}
=
{}^{W}\widehat T_{C_{n,k}}
\left({}^{B}T_C\right)^{-1}.
\label{eq:body_transform}
\end{equation}
This conversion changes the represented rigid body, not the world reference frame. Replacing ${}^{W}T_{C_n}$ with an activation-time pose would incorrectly translate and rotate the world-anchored proposal. The resulting body-pose sequence remains a motion reference, not a directly executable vehicle-attitude command.

% 中文翻译：拍摄时刻的坐标落地。轨迹提议以 t_n 时刻拍摄的图像为条件，而非激活时的飞行器状态。首先使用拍摄时刻的相机世界位姿，将以该相机坐标系表示的预测位姿转换到世界系。若 B_T_C 表示相机坐标到机体坐标的刚体变换，则再右乘其逆矩阵，得到世界坐标系下的机体位姿参考。这里改变的是所描述的刚体，而不是世界参考坐标系。用激活时的位姿替换拍摄位姿会错误地平移和旋转整条世界系提议。转换后的机体位姿序列仍是运动参考，不是可以直接执行的飞行器姿态指令。

\paragraph{Scheduled-state prediction.}
Let $s(t_{s,n})$ denote the latest state estimate used for planning, with timestamp $t_{s,n}\leq t_{h,n}$, and let $\tau_n^{-}$ denote the currently committed world-frame reference, parameterized by absolute time. The prospective activation state is
\begin{equation}
\widetilde s_{h,n}
=
\Phi\!\left(
s(t_{s,n}),
\tau_n^{-},
t_{h,n}-t_{s,n}
\right),
\label{eq:prospective_state}
\end{equation}
where $\Phi$ propagates the state under the committed motion. Its propagation interval starts at the state-estimation timestamp, not automatically at the image-capture time. The planner uses this prediction together with the committed reference to assess the transition at $t_{h,n}$. Prediction error and reference-tracking error are not assumed to be zero.

% 中文翻译：预定交接状态预测。令 $s(t_{s,n})$ 为规划所用的最新状态估计，其时间戳满足 $t_{s,n}\leq t_{h,n}$；令 $\tau_n^{-}$ 为按绝对时间参数化的当前世界系参考轨迹。$\Phi$ 根据当前已提交运动将状态从 $t_{s,n}$ 传播到预定交接时刻。传播时长从状态估计时间戳开始计算，而不是直接从图像拍摄时刻开始。规划器结合该预测和当前参考轨迹评估交接可行性，不假设状态预测误差或参考跟踪误差为零。

The controller must check the latest state and plan validity before activation. The actual activation time is recorded separately from $t_{h,n}$. If timing jitter, a changed committed plan, or state mismatch invalidates the planned transition, the proposal must be realigned and revalidated or rejected; changing its timestamp alone is insufficient.

% 中文翻译：控制器在激活前必须核查最新状态与计划有效性，并单独记录实际激活时间。若时间抖动、已提交计划变化或状态失配使过渡失效，必须重新对齐并校验或者拒绝该提议；仅修改时间戳不足以解决问题。

\paragraph{Progress-aware transition requirements.}
Elapsed preparation time is not equivalent to progress along the newly predicted trajectory, because the UAV has been following $\tau_n^{-}$ rather than the new proposal. The execution interface therefore does not remove a proposal prefix solely according to $\Delta_n$. A prefix may be omitted only when the executed motion and task context establish that the corresponding requirement has already been satisfied. For repeated or self-intersecting motions, geometric proximity alone is insufficient to determine task progress.

% 中文翻译：进度感知的过渡要求。准备过程已经过去的时间，不等同于新预测轨迹上的执行进度，因为无人机一直在跟随旧轨迹，而非新提议。因此，执行接口不能仅按照观测到交接的延迟删除新轨迹前缀。只有实际执行运动和任务上下文能够确认相关要求已经完成时，才允许略去该前缀。对于重复或自交运动，仅凭空间距离近并不足以确定任务进度。

The transition must connect the scheduled boundary state to a compatible part of the remaining proposal without skipping uncompleted task requirements. World-frame targets and the order of required maneuvers are retained; preserving only the endpoint is insufficient for passage, orbiting, or side-specific motion. The predicted sample times specify nominal motion timing, whereas the downstream planner determines feasible execution timing. Any retiming must preserve explicit duration, speed, and ordering constraints in the instruction. If progress cannot be established or no task-consistent feasible connection exists, the proposal must be rejected or refreshed instead of being forced into the current execution state.

% 中文翻译：过渡轨迹必须把预定边界状态连接到剩余提议中相容的部分，并且不能跳过未完成的任务要求。世界系目标和规定动作的先后顺序应被保留；对于穿越、环绕或指定侧向的运动，仅保留终点是不够的。预测采样时刻表示名义运动时序，可行执行时序由下游规划器确定，但重新定时不能破坏指令明确规定的持续时间、速度和顺序约束。若无法确认进度，或不存在兼顾任务语义与动力学的连接，应拒绝或刷新提议，而非强行适配当前状态。

\paragraph{Reference continuity and acceptance.}
Let $\tau_n^{+}(\xi)$ denote the replacement reference with local execution time $\xi=t-t_{h,n}$. At the scheduled handoff, the minimum kinematic reference-continuity conditions considered here are
\begin{equation}
p_n^{+}(0)=p_n^{-}(t_{h,n}),
\qquad
v_n^{+}(0)=v_n^{-}(t_{h,n}).
\label{eq:handoff_continuity}
\end{equation}
All positions and velocities are expressed in the world frame. These equalities constrain commanded references; they do not assert that the actual vehicle state exactly equals the nominal reference. The predicted and latest estimated states must remain compatible with the tracking conditions accepted by the flight stack. If a separate transition segment is followed by a retained proposal segment, continuity must also hold at their connection, not only at activation.

% 中文翻译：参考连续性与接受条件。新参考轨迹使用以交接时刻为零点的局部执行时间，旧参考使用绝对时间。交接点至少满足世界坐标系下的位置和速度连续。该等式约束的是指令参考轨迹，而不是断言真实飞行器状态与名义参考完全相等。预测状态与最新估计状态仍需符合飞行系统允许的跟踪条件。如果先执行独立过渡段再接保留的提议段，两段之间也必须满足连续性，而不只是激活点连续。

\section{Results}
% 中文翻译：实验结果
\label{sec:results}

To comprehensively evaluate DiffWAM, we conduct extensive benchmark, simulation, real-world, deployment, and ablation experiments focusing on four key questions: (1) How does DiffWAM compare with representative world-action and trajectory-prediction methods across diverse aerial navigation tasks? (2) Can the predicted motion be reliably executed in simulation and real-world environments, including tasks that require continuous and spatially structured trajectories? (3) Can the proposed efficient inference and asynchronous execution pipeline support practical onboard UAV deployment? (4) Which representation, predictive-computation, geometric, and training choices contribute to trajectory quality and execution performance?
% 中文翻译：为全面评估 DiffWAM，我们开展了广泛的基准、仿真、真实世界、部署和消融实验，重点关注以下四个关键问题：（1）在多样化的空中导航任务中，DiffWAM 与具有代表性的世界动作方法和轨迹预测方法相比表现如何？（2）预测的运动能否在仿真和真实世界环境中得到可靠执行，包括那些需要连续且具有空间结构的轨迹的任务？（3）所提出的高效推理与异步执行流水线能否支持实际的无人机端侧部署？（4）哪些表征、预测计算、几何和训练方面的选择有助于提升轨迹质量和执行性能？

To address these questions, Sec.~\ref{sec:exp_setup} first introduces the training datasets, evaluation datasets, evaluation metrics, baselines, and experimental protocol. Sec.~\ref{sec:benchmark_results} then presents quantitative benchmark comparisons and simulation results. Sec.~\ref{sec:real_results} evaluates real-world navigation and onboard deployment. Finally, Sec.~\ref{sec:training_results} provides systematic ablation studies to analyze the major design choices of DiffWAM.
% 中文翻译：为回答这些问题，Sec.~\ref{sec} 首先介绍训练数据集、评估数据集、评估指标、基线方法和实验协议。随后，Sec.~\ref{sec} 给出定量基准比较和仿真结果。Sec.~\ref{sec} 进一步评估真实世界导航和端侧部署。最后，Sec.~\ref{sec} 通过系统的消融实验分析 DiffWAM 的主要设计选择。

% ================================================================
\subsection{Experimental Setup}
% 中文翻译：实验设置
\label{sec:exp_setup}

\subsubsection{Training Datasets}
% 中文翻译：训练数据集

To train DiffWAM, we construct a large-scale aerial navigation dataset primarily using synthetic data generated by two automated pipelines developed in our previous work, FlyMirage~\cite{flymirage} and NavGen~\cite{huang2026navgen}. FlyMirage first employs large language models (LLMs) to generate diverse scene descriptions and uses the generative model Marble~\cite{worldlabs2025marble} to construct corresponding high-fidelity 3DGS environments. A heuristic autonomous exploration strategy is then applied to traverse the generated scenes, while Boxer~\cite{boxer2026} detects and annotates object categories and their 3D locations. Based on these annotations, the efficient trajectory planner GCOPTER~\cite{shao2024design} automatically generates dynamically feasible UAV trajectories between objects and spatial regions. The resulting FlyMirage dataset contains approximately 1,100 scenes and more than 110K trajectories, alleviating several common limitations of existing aerial navigation datasets, including inconsistent observation quality, limited scene diversity, and trajectories that do not satisfy realistic UAV dynamics. In parallel, NavGen provides a complementary data-generation paradigm based on generative video models. By exploiting the rich visual and motion priors encoded in video foundation models, NavGen efficiently generates task-conditioned navigation trajectories for diverse instructions, such as ``follow the path and continue moving forward,'' while data augmentation further improves the diversity of scene appearance and environmental configurations. More importantly, the strong generative and motion generalization capabilities of video models enable NavGen to synthesize complex motion patterns that are difficult to obtain using conventional point-to-point navigation pipelines, such as ``fly through the cave'' and ``complete one orbit around the cabin in the forest.'' In this way, NavGen substantially expands the training distribution from goal-directed navigation to continuous motions with richer spatial and geometric structures, yielding approximately 400K trajectories covering a broad range of aerial navigation tasks. We have released an initial batch of the data in these projects.
% 中文翻译：为了训练 DiffWAM，我们构建了一个大规模空中导航数据集，主要使用由我们此前工作中开发的两套自动化管线 FlyMirage~\cite{flymirage} 和 NavGen~\cite{navgen} 生成的合成数据。FlyMirage 首先利用大语言模型（LLMs）生成多样化的场景描述，并使用生成模型 Marble~\cite{?} 构建相应的高保真 3DGS 环境。随后，采用启发式自主探索策略遍历所生成的场景，同时由 Boxer~\cite{?} 检测并标注物体类别及其三维位置。基于这些标注，高效轨迹规划器 GCOPTER~\cite{?} 自动生成物体与空间区域之间满足动力学可行性的无人机轨迹。最终得到的 FlyMirage 数据集包含约 1,100 个场景和超过 110K 条轨迹，缓解了现有空中导航数据集中若干常见的局限，包括观测质量不一致、场景多样性有限，以及轨迹不满足真实无人机动力学要求等问题。

% 与此同时，NavGen 提供了一种基于生成式视频模型的互补数据生成范式。通过利用视频基础模型中编码的丰富视觉与运动先验，NavGen 能够针对多样化指令高效生成任务条件化的导航轨迹，例如“沿着路径前进并继续向前运动”，同时通过数据增强进一步提升场景外观和环境配置的多样性。更重要的是，视频模型强大的生成与运动泛化能力使 NavGen 能够合成使用传统点到点导航管线难以获得的复杂运动模式，例如“飞过洞穴”和“绕森林中的小木屋完成一圈环绕”。通过这种方式，NavGen 将训练分布从目标导向导航显著扩展到具有更丰富空间与几何结构的连续运动，最终获得约 400K 条覆盖广泛空中导航任务的轨迹。我们已经先期在这些项目中开源了一批数据。

% In addition to synthetic data, we incorporate real-world flight data collected from extensive physical UAV exploration and navigation experiments. The raw flight data are systematically cleaned and filtered according to visual quality, trajectory validity, sensor integrity, and instruction--trajectory consistency, while invalid, incomplete, or low-quality samples are removed before training. These real-world samples complement synthetic data with realistic visual appearance, sensor characteristics, motion perturbations, and environmental variations, further improving the coverage of real deployment conditions.
% % 中文翻译：除合成数据外，我们还引入了来自大量真实无人机探索和导航实验的实飞数据。原始飞行数据按照视觉质量、轨迹有效性、传感器数据完整性以及指令--轨迹一致性进行系统清洗与筛选，并在训练前去除无效、不完整或质量较低的样本。这些真实飞行数据能够补充合成数据中难以完整建模的真实视觉外观、传感器特性、运动扰动和环境变化，从而进一步提升训练数据对实际部署场景的覆盖能力。

Combining the above methods, we obtain more than 510K training samples spanning a broad range of aerial navigation capabilities, including \emph{Basic Motion}, \emph{Instruction Following}, \emph{Object Navigation}, \emph{Precise Navigation}, \emph{Spatial Grounding}, \emph{Specific Trajectory}, \emph{Language Control}, \emph{Scene Understanding}, and \emph{Object Searching}. This heterogeneous data distribution exposes DiffWAM to both elementary motion primitives and complex language-conditioned trajectory structures, providing a diverse data foundation for learning continuous and generalizable UAV navigation.
% 中文翻译：综合上述方法，我们获得了超过 510K 条训练样本，覆盖广泛的空中导航能力，包括基础运动（Basic Motion）、指令跟随（Instruction Following）、目标导航（Object Navigation）、精确导航（Precise Navigation）、空间定位（Spatial Grounding）、特定轨迹（Specific Trajectory）、语言控制（Language Control）、场景理解（Scene Understanding）和目标搜索Object Searching）。这种异构的数据分布使 DiffWAM 同时接触到基础运动原语和复杂的语言条件化轨迹结构，为学习连续且具有泛化能力的无人机导航提供了多样化的数据基础。

\subsubsection{Evaluation Datasets}
% 中文翻译：评估数据集

\begin{figure}[h]
    \centering
    \includegraphics[width=\linewidth]
    {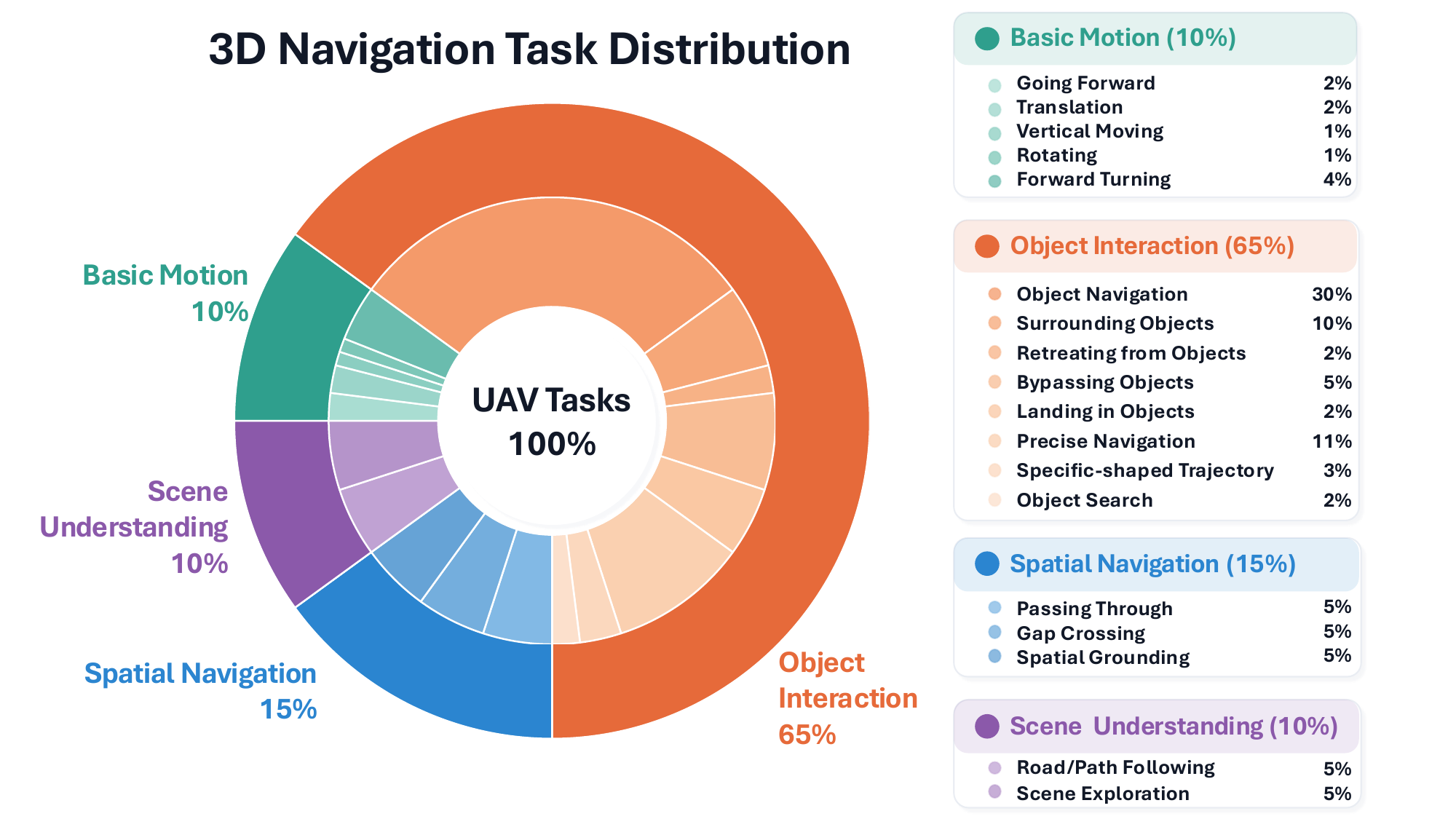}
    \caption{\textbf{Distribution of the evaluation tasks.}
    The benchmark contains 18 task types grouped into four families:
    basic motion, object interaction, spatial navigation, and
    scene understanding. Percentages denote the fraction of the
    complete evaluation set.}
    % 中文翻译：评估任务分布。基准包含18种任务，划分为基础飞行控制、物体交互、空间穿越和环境导航四类。百分比表示各任务在完整评估集中的占比。
    \label{fig:dataset_distribution}
\end{figure}

We evaluate DiffWAM on a unified aerial-navigation benchmark covering both simulation and real-world scenes. The benchmark contains 1,000 test samples, which we refer to as \textbf{DiffWAM-1000}, and fully represents the distribution of all task categories. It covers 18 task types organized into four families, as illustrated in Fig.~\ref{fig:dataset_distribution}: basic motion (10\%), object interaction (65\%), spatial navigation (15\%), and scene understanding (10\%).
% 中文翻译：我们在一个统一的空中导航基准上评估 DiffWAM，该基准同时涵盖仿真场景和真实世界场景。如 Fig.~\ref{fig} 所示，评估集包含 18 种任务类型，并被组织为四个类别：基础运动（Basic Motion，10%）、物体交互（Object Interaction，65%）、空间导航（Spatial Navigation，15%）和场景理解（Scene Understanding，10%）。图中还完整展示了各项任务所占的比例。

Tab.~\ref{tab:tasks500} provides a more detailed description of each task category of the test benchmark, as well as examples of typical task instructions. The benchmark is designed to evaluate complementary aspects of continuous UAV navigation. Basic motion tasks examine whether the predicted trajectory follows explicit translational and rotational commands. Object-interaction tasks require the model to associate language instructions with target objects and generate target-relative motion, including approaching, retreating, side-specific passing, orbiting, slalom flight, and landing. Spatial-navigation tasks evaluate motion through constrained free-space regions, while scene-understanding tasks require the predicted motion to follow larger-scale scene structures such as roads or boundaries. Object-interaction tasks constitute the largest subset because language-conditioned target selection and spatial-relation reasoning are central to vision-language UAV navigation.
% 中文翻译：Tab.~\ref{tab} 对测试基准中的各个任务类别进行了更详细的描述，并给出了典型任务指令的示例。该基准旨在评估连续无人机导航中相互补充的不同方面。基础运动任务考察预测轨迹是否遵循明确的平移和旋转指令。物体交互任务要求模型将语言指令与目标物体相关联，并生成相对于目标的运动，包括接近、远离、指定侧通过、环绕、蛇形飞行和降落。空间导航任务评估无人机在受限自由空间区域中的运动，而场景理解任务则要求预测运动遵循道路或边界等更大尺度的场景结构。物体交互任务构成最大的任务子集，因为语言条件下的目标选择和空间关系推理是视觉语言无人机导航的核心。

\begin{table}[h]
\centering
\small
\renewcommand{\arraystretch}{1.2}
\caption{\textbf{Composition of the evaluation benchmark.} All task percentages are computed with respect to the complete evaluation set.}
% 中文翻译：评估基准的任务组成。所有任务比例均相对于完整评估集计算。
\label{tab:tasks500}

\begin{tabularx}{\linewidth}{llX}
\toprule
\textbf{Task family} & \textbf{Task type} & \textbf{Example of instructions} \\
\midrule
Basic Motion
    & Going Forward  & Fly forward for 5 seconds \\
    & Translation & Move left without turning \\
    & Vertical Moving & Move upwards by 3 meters \\
    & Rotating  & Rotate 45 degrees in place \\
    & Forward Turning & Move to the right and face that direction  \\ \cline{2-3}

Object Interaction
    & Object Navigation & Navigate to the black rock formation \\
    & Surrounding objects  & Circle around the white pillar \\
    & Retreating from objects & Step back and leave the square box \\
    & Bypassing objects  & Go around from the right side of the tree \\
    & Landing in Objects & Land on the brown table \\
    & Precise Navigation & Navigate to the second billboard on the left \\
    & Specific-shaped trajectory & Fly in an S-shaped trajectory \\
    & Object Search & Find a place where one can drink water \\ \cline{2-3}

Spatial navigation
    & Passing Through  & Pass through the middle of the two charging stations \\
    & Gap Crossing  & Go through the opening in this window \\
    & Spatial Grounding & Head to the left of the chair on the right \\ \cline{2-3}

Scene Understanding
    & Road/path Following & Keep flying along the path in the forest \\
    & Scene exploration & Keep exploring along the corridor of the room until you reach the end of the corridor \\
\bottomrule
\end{tabularx}
\end{table}

Training, validation, and evaluation data are separated at the scene level to prevent visually adjacent viewpoints or paraphrased instructions from appearing in different partitions. Unless otherwise specified, all benchmark results in Secs.~\ref{sec:benchmark_results} and~\ref{sec:real_results} are reported on this revised evaluation protocol.
% 中文翻译：训练集、验证集和测试集按照场景进行隔离，避免视觉上相邻的视角或同一指令的不同改写版本出现在不同数据划分中。除非特别说明，Secs.~\ref{sec:benchmark_results} 和~\ref{sec:real_results} 中的全部基准结果均采用该修订后的统一评估协议。

% ----------------------------------------------------------------
\subsubsection{Evaluation Metrics}
% 中文翻译：评估指标

We evaluate the proposed system from five complementary perspectives: endpoint accuracy, trajectory accuracy, rotation accuracy, closed-loop task completion, and execution efficiency.
% 中文翻译：我们从三个互补维度评估所提出的系统：终点精度、完整轨迹质量以及闭环任务完成情况。

\paragraph{Endpoint accuracy.}
Endpoint accuracy evaluates whether the UAV reaches sufficiently close to the desired target and is mainly used for tasks such as object navigation and instruction following. For the $i$-th test episode, we denote the ground-truth trajectory as
$\tau_i^{\star}=\{P_{i,1}^{\star},P_{i,2}^{\star},\ldots,P_{i,N}^{\star}\}$,
and the trajectory predicted by DiffWAM as
$\hat{\tau}_i=\{\hat{P}_{i,1},\hat{P}_{i,2},\ldots,\hat{P}_{i,N}\}$,
where $P_{i,n}^{\star}$ and $\hat{P}_{i,n}$ denote the ground-truth and predicted 3D positions at the $n$-th trajectory waypoint, respectively, and $N$ is the number of temporally aligned trajectory waypoints. We measure the endpoint deviation using the Final Displacement Error (FDE):
% 中文翻译：终点精度：终点精度用于衡量无人机是否能够最终到达期望目标附近，主要用于目标导航和指令跟随等任务。对于第 $i$ 个测试任务，我们将其真值轨迹记为 $\tau_i^{\star}=\{P_{i,1}^{\star},P_{i,2}^{\star},\ldots,P_{i,N}^{\star}\}$，将 DiffWAM 预测的轨迹记为 $\hat{\tau}_i=\{\hat{P}_{i,1},\hat{P}_{i,2},\ldots,\hat{P}_{i,N}\}$，其中 $P_{i,n}^{\star}$ 和 $\hat{P}_{i,n}$ 分别表示第 $n$ 个轨迹航点的真值三维位置和预测三维位置，$N$ 表示经过时间对齐后的轨迹航点数量。我们采用终点位移误差（Final Displacement Error, FDE）衡量预测轨迹与真值轨迹在终点处的偏差：

\begin{equation}
\mathrm{FDE}_i
=
\left\|
\hat{P}_{i,N}
-
P_{i,N}^{\star}
\right\|_2 .
\label{eq:fde}
\end{equation}

An episode is considered successful if its endpoint error satisfies $\mathrm{FDE}_i \leq 1$~m in indoor environments or $\mathrm{FDE}_i \leq 3$~m in outdoor environments. Endpoint accuracy is then reported as the percentage of successful episodes over all evaluated trials. Invalid predictions, timeouts, and episodes whose endpoint errors exceed the corresponding distance threshold are counted as failures.
% 中文翻译：如果一个测试回合的终点误差在室内环境中满足 $\mathrm{FDE}_i \leq 1$~m，或在室外环境中满足 $\mathrm{FDE}_i \leq 3$~m，则该回合被视为成功。随后，终点准确率被报告为所有评估试验中成功回合所占的百分比。无效预测、超时以及终点误差超过相应距离阈值的回合均被计为失败。

\paragraph{Trajectory accuracy.}
Endpoint accuracy alone is insufficient to characterize the geometric quality of a continuous trajectory. For example, in an orbiting task, the UAV may start and terminate at nearly the same position, resulting in a small endpoint error even if the intermediate motion does not follow the required circular path around the target. Therefore, in addition to endpoint accuracy, we evaluate the entire predicted trajectory using root-mean-square error (RMSE), average displacement error (ADE), and mean orientation error. For the $i$-th test episode containing $N$ temporally aligned waypoints, RMSE and ADE are defined as
% 中文翻译：轨迹精度。仅靠终点精度不足以刻画连续轨迹的几何质量。例如，在环绕任务中，无人机可能从几乎相同的位置开始并结束，因此即使中间运动没有遵循绕目标所要求的圆形路径，也可能产生较小的终点误差。因此，除终点精度外，我们还使用均方根误差（RMSE）、平均位移误差（ADE）和平均朝向误差对完整的预测轨迹进行评估。对于包含 $N$ 个时间对齐航点的第 $i$ 个测试回合，RMSE 和 ADE 定义为：
\begin{equation}
\mathrm{RMSE}_i
=
\sqrt{
\frac{1}{N}
\sum_{n=1}^{N}
\left\|
\tilde{P}_{i,n}
-
P^{\star}_{i,n}
\right\|_2^2
},
\qquad
\mathrm{ADE}_i
=
\frac{1}{N}
\sum_{n=1}^{N}
\left\|
\tilde{P}_{i,n}
-
P^{\star}_{i,n}
\right\|_2 .
\label{eq:trajectory_error}
\end{equation}

Here, $P^{\star}_{i,n}$ and $\tilde{P}_{i,n}$ denote the ground-truth and DiffWAM-predicted 3D positions at the $n$-th waypoint, respectively. RMSE emphasizes larger trajectory deviations, whereas ADE measures the average positional deviation over the entire trajectory. The mean orientation error evaluates the discrepancy between the predicted and reference orientations along the trajectory. All trajectory errors are computed in the initial camera coordinate frame without post-hoc rigid transformation or scale alignment, and the shared initial pose is excluded from the error computation.
% 中文翻译：其中，$P^{\star}{i,n}$ 和 $\tilde{P}{i,n}$ 分别表示第 $n$ 个轨迹航点处的真值三维位置和 DiffWAM 预测的三维位置。RMSE 更强调较大的轨迹偏差，而 ADE 衡量整条轨迹上的平均位置偏差。平均朝向误差用于评估沿轨迹的预测朝向与参考朝向之间的差异。所有轨迹误差均在初始相机坐标系下计算，不进行事后的刚体变换或尺度对齐，并且共享的初始位姿不参与误差计算。

\paragraph{Rotation accuracy.}
Let $R^{\star}_{i,n},\tilde{R}_{i,n}\in\mathrm{SO}(3)$ denote
the reference and predicted orientations at waypoint $n$,
expressed relative to the initial camera frame.
The mean rotation error in degrees is
\begin{equation}
\mathrm{RoE}_i
= \frac{180}{\pi N}\sum_{n=1}^{N}
\arccos\!\left(
\operatorname{clip}\!\left(
\frac{\operatorname{tr}\!\left(
(R^{\star}_{i,n})^{\top}\tilde{R}_{i,n}
\right)-1}{2},-1,1
\right)\right).
\label{eq:orientation_error}
\end{equation}
% 它衡量每个航点完整三维朝向的角度误差，再沿轨迹取平均。相对旋转预测需先逐步组合为相对于初始帧的姿态；不是单步旋转误差，也不只是 yaw。排除共享初始位姿，最后对各测试回合等权平均。

\paragraph{Closed-loop task success.}
For simulation and real-world execution, we further report task success rate (SR). Unlike endpoint accuracy, SR evaluates whether the complete semantic motion requirement has been satisfied. For goal-reaching tasks, success requires reaching the target within the prescribed positional tolerance without collision or manual intervention. For orbiting, side-specific passage, gap traversal, landing, and other structured-motion tasks, additional task-specific conditions such as traversal direction, angular coverage, heading, clearance, and landing/contact status are evaluated. Timeouts, collisions, invalid commands, and incomplete maneuvers are counted as failures.
% 中文翻译：闭环任务成功率。对于仿真和真实世界执行，我们进一步报告任务成功率（SR）。与终点准确率不同，SR 用于评估完整的语义运动要求是否得到满足。对于目标到达任务，成功要求无人机在无碰撞或人工干预的情况下，到达规定的位置容差范围内。对于环绕、指定侧通过、间隙穿越、降落以及其他结构化运动任务，还会评估通过方向、角度覆盖、航向、净空以及降落/接触状态等额外的任务特定条件。航向容差设置为 xxx 度。超时、碰撞、无效指令以及未完成的机动动作均计为失败。

\paragraph{Execution efficiency.}
For deployment experiments, we report model inference latency using the median (P50) and 95th percentile (P95), peak memory usage, end-to-end navigation-update latency, task completion time and waiting fraction. Latency measurements exclude model loading unless otherwise stated. All hardware, numerical precision, caching, and distributed-inference configurations are reported together with their corresponding accuracy measurements.
% 中文翻译：执行效率。对于部署实验，我们报告模型推理延迟的中位数（P50）和第 95 百分位数（P95）、峰值内存占用、端到端导航更新延迟、任务完成时间、等待时间占比以及轨迹交接不连续性。除非另有说明，延迟测量不包括模型加载时间。所有硬件、数值精度、缓存和分布式推理配置均与其对应的准确率测量结果一起报告。

% ----------------------------------------------------------------
\subsubsection{Baselines and Evaluation Protocol}
% 中文翻译：对比方法与评估协议

We distinguish overall navigation-benchmark comparisons from supplementary trajectory-readout comparisons. For the overall evaluation on IndoorUAV-VLA~\cite{indooruav}, UAV-FLOW-Sim~\cite{uavflow}, and DiffWAM-1000, we compare DiffWAM with WorldVLN~\cite{worldvln}, ImagineUAV~\cite{imagineuav}, Fast-WAM-UAV~\cite{fastwam}, and WorldFly~\cite{worldfly}. These comparisons assess endpoint performance across the three datasets, with additional task-family results reported for DiffWAM-1000.
% 中文翻译：我们区分完整导航基准对比与补充性的轨迹读出对比。在IndoorUAV-VLA、UAV-FLOW-Sim和DiffWAM-1000上的完整评估中，我们将DiffWAM与WorldVLN、ImagineUAV、Fast-WAM-UAV和WorldFly进行比较。这些对比评估三个数据集上的终点性能，并进一步报告DiffWAM-1000各任务族的结果。

Within each comparison, methods receive the same evaluation instructions and observations whenever their respective interfaces permit. Training, validation, and evaluation partitions are separated at the scene level. All entries in the overall benchmark comparison use the common endpoint criterion. This endpoint-based success rate is distinguished from task-specific closed-loop completion: reaching the reference endpoint alone does not establish that the required traversal direction, orbit coverage, intermediate motion, or landing condition has been satisfied.
% 中文翻译：在每组对比中，各方法在其接口允许的条件下接收相同的评估指令与观测。训练集、验证集和评估集按场景划分。完整基准对比中的所有结果均采用统一的FDE不超过1米的终点判据。该终点成功率与任务特定的闭环完成率相区分：仅到达参考终点，并不代表已经满足指令要求的穿越方向、环绕覆盖范围、中间运动过程或降落条件。

% ================================================================
\subsection{Benchmark Results and Analysis}
% 中文翻译：基准实验结果与分析
\label{sec:benchmark_results}

\begin{figure}[!h]
\centering
\includegraphics[
width=\linewidth,
height=\textheight,
keepaspectratio,
clip]
{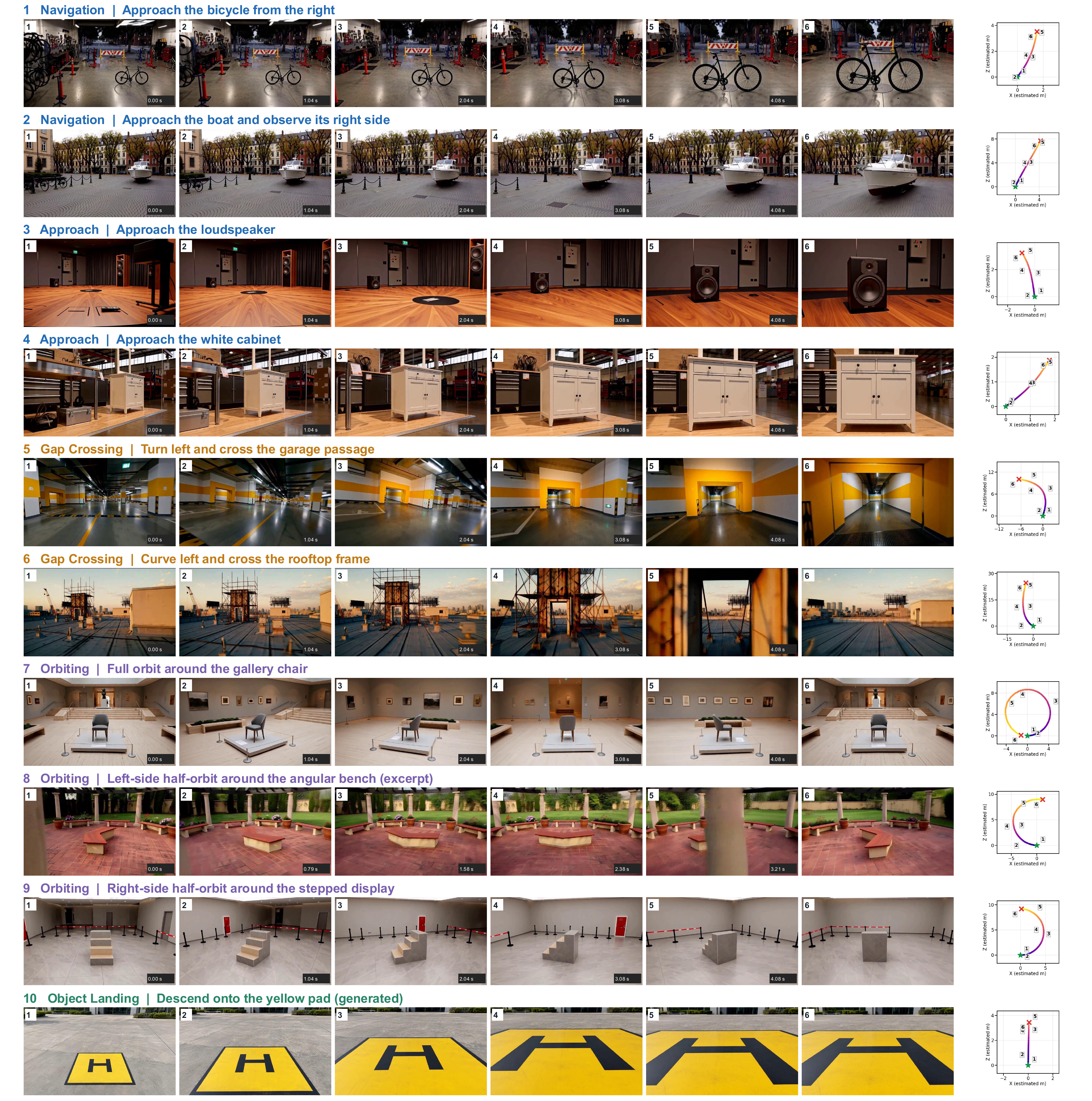}
\caption{\textbf{Representative trajectory-generation results of DiffWAM.}
Rows 1--4 illustrate target-directed navigation and approach; Rows 5--6 show constrained gap traversal; and Rows 7--9 show a full orbit and direction-conditioned half-orbits, including the excerpt in Row 8. Row 10 provides a supplemental generated example of descent onto a yellow landing pad. Each visual sequence is accompanied by a top-view trajectory. These selected examples support qualitative analysis rather than aggregate success-rate estimation.}
% 中文翻译：DiffWAM 的代表性轨迹生成结果。第1--4行展示目标导向导航与接近，第5--6行展示受限间隙穿越，第7--9行展示完整环绕和方向条件化半环绕，其中第8行为节选。前九行为代表性的 DiffWAM-1000 案例，第10行为下降至黄色降落垫的补充生成案例。每组视觉序列均附有俯视轨迹。这些选取的案例用于定性分析，而非估计总体成功率。
\label{fig:sim_results}
\end{figure}

\subsubsection{DiffWAM Performance}
% 中文翻译：DiffWAM性能

We assess DiffWAM through the endpoint benchmark in Table~\ref{tab:accuracy} and the trajectory examples in Fig.~\ref{fig:sim_results}. On DiffWAM-1000, DiffWAM achieves a benchmark success rate of 74.40\% under the shared endpoint criterion. The qualitative examples complement this endpoint-based measurement by showing how the predicted trajectory changes with the motion requirements of the instruction.
% 中文翻译：我们通过 Table~\ref{tab:accuracy} 中的终点基准和 Fig.~\ref{fig:sim_results} 中的轨迹案例评估 DiffWAM。在 DiffWAM-1000 上，DiffWAM 在统一的一米终点判据下取得74.40\%的基准成功率。定性案例进一步展示了预测轨迹如何随指令中的运动要求变化，从而补充终点指标所提供的信息。

For target-directed navigation and approach, Rows 1--4 of Fig.~\ref{fig:sim_results} illustrate both object selection and target-relative motion. The bicycle case requires approaching from the right, whereas the boat case requires approaching the boat and observing its right side. The loudspeaker and white-cabinet cases illustrate direct approach in different scene layouts. The visual sequences and top-view paths show progressive motion toward the referenced objects, with lateral displacement adapted to the target-relative instruction and initial viewpoint.
% 中文翻译：对于目标导向导航和接近任务，Fig.~\ref{fig:sim_results} 的第1--4行展示了目标选择以及相对于目标的运动。自行车案例要求从右侧接近，船只案例则要求接近船只并观察其右侧。音箱和白色柜子案例展示了不同场景布局中的直接接近。视觉序列与俯视路径呈现出朝向所指目标逐步推进的运动，同时根据目标相对指令和初始视角调整横向位移。

Rows 5--6 illustrate constrained traversal. In the garage case, the trajectory turns left before passing through the passage; in the rooftop case, it curves left toward and through the frame. These examples involve a sequence of approach, alignment, and traversal rather than simply terminating near an opening. They therefore highlight a geometric requirement that endpoint proximity alone cannot fully characterize.
% 中文翻译：第5--6行展示受限空间穿越。在车库案例中，轨迹先向左转弯，再穿过通道；在屋顶案例中，轨迹向左弯曲并穿过框架。这些案例包含接近、对齐和穿越的连续过程，而不是仅停留在开口附近，因此体现了终点距离难以完整刻画的几何要求。

The orbiting cases in Rows 7--9 exhibit more structured motion. Row 7 shows a full orbit around the gallery chair, Row 8 shows an excerpt of a left-side half-orbit around the angular bench, and Row 9 shows a right-side half-orbit around the stepped display. Their top-view paths contain a loop or direction-dependent arcs, consistent with the corresponding instructions. Unlike point-goal navigation, these tasks require the intermediate path to evolve around a reference object. For a full orbit, the start and end positions may be close even when the intervening motion is incorrect; consequently, endpoint error must be complemented by full-trajectory evaluation when assessing such behavior.
% 中文翻译：第7--9行的环绕案例呈现出更强的运动结构。第7行展示围绕展厅椅子的完整环绕，第8行展示围绕转角长椅的左侧半环绕节选，第9行展示围绕阶梯展台的右侧半环绕。相应俯视路径呈现闭环或具有方向差异的弧线，与对应指令一致。不同于点目标导航，这些任务要求中间路径围绕参照物持续变化。对于完整环绕，即使中间运动错误，起终点也可能十分接近，因此评估此类行为时需要在终点误差之外考察完整轨迹。

Row 10 provides a supplemental landing example. As the viewpoint approaches the yellow pad, the pad occupies an increasing portion of the image, illustrating the requested descent toward the landing region. This example extends the visualization beyond horizontal approach and orbiting, but does not provide an independently measured landing error or a physical landing success rate.
% 中文翻译：第10行提供了一个补充生成的降落案例。随着视点接近黄色垫子，垫子在图像中所占比例逐渐增大，展示了朝向降落区域下降的指令效果。这一案例扩展了水平接近和环绕之外的可视化内容，但并未提供独立测量的降落误差或真实降落成功率。

Taken together, these examples show task-dependent differences in trajectory geometry: relatively direct target approach, turning motion through constrained openings, circumferential motion around objects, and a generated descent sequence. They illustrate that the model output encodes how to move relative to the scene, rather than only where to terminate. The benchmark results below quantify endpoint performance separately from these selected qualitative demonstrations.
% 中文翻译：总体而言，这些案例展示了随任务变化的轨迹几何特征，包括相对直接的目标接近、穿越受限开口的转向运动、围绕物体的周向运动以及生成的下降序列。它们表明模型输出不仅描述运动终点，也表达相对于场景如何运动。下文的基准结果对终点性能进行独立量化，不将这些选取的定性案例当作总体成功率统计。

\subsubsection{Overall Benchmark Performance}
% 中文翻译：整体基准性能

Table~\ref{tab:accuracy} compares DiffWAM with WorldVLN, ImagineUAV, Fast-WAM-UAV, and WorldFly on IndoorUAV-VLA, UAV-FLOW-Sim, and DiffWAM-1000. Under the common endpoint criterion, DiffWAM achieves success rates of 56.77\%, 91.42\%, and 74.40\%, respectively, ranking first on all three benchmarks among the compared methods. WorldVLN is the strongest competing method in each overall comparison, with 39.32\%, 80.24\%, and 58.40\%, respectively. The corresponding absolute improvements are 17.45, 11.18, and 16.00 percentage points.
% 中文翻译：Table~\ref{tab:accuracy} 在 IndoorUAV-VLA、UAV-FLOW-Sim 和 DiffWAM-1000 上比较了 DiffWAM、WorldVLN、ImagineUAV、Fast-WAM-UAV 和 WorldFly。在统一的一米终点判据下，DiffWAM 分别取得56.77\%、91.42\%和74.40\%的成功率，在三个基准的对比方法中均排名第一。WorldVLN 在三个基准上均为总体表现最强的对比方法，对应结果分别为39.32\%、80.24\%和58.40\%。DiffWAM 的绝对提升分别为17.45、11.18和16.00个百分点。

\begin{table}[!h]
\centering
\small
\caption{\textbf{Benchmark success rate (SR, \%) on IndoorUAV-VLA, UAV-FLOW-Sim, and DiffWAM-1000.} All reported entries use the same endpoint criterion; this endpoint-based SR is distinct from task-specific closed-loop completion. BM: basic motion; OI: object interaction; SN: spatial navigation; SU: scene understanding.}
% 中文翻译：IndoorUAV-VLA、UAV-FLOW-Sim 和 DiffWAM-1000 上的基准成功率（SR，\%）。所有结果均使用相同的一米终点判据；该终点成功率不同于按具体运动要求评估的闭环任务完成率。BM：基础运动；OI：物体交互；SN：空间导航；SU：场景理解。
\label{tab:accuracy}

\begin{tabularx}{\linewidth}{Xrrrrrrr}
\toprule
& \multicolumn{1}{c}{\textbf{IndoorUAV-VLA}}
& \multicolumn{1}{c}{\textbf{UAV-FLOW-Sim}}
& \multicolumn{5}{c}{\textbf{DiffWAM-1000}} \\
\cmidrule(lr){2-2}
\cmidrule(lr){3-3}
\cmidrule(lr){4-8}
\multicolumn{1}{X}{\textbf{Method}}
& Average & Average & BM & OI & SN & SU & Average \\
\midrule
WorldVLN~\cite{worldvln}
& 39.32 & 80.24 & 86.00 & 56.62 & 50.77 & 54.00 & 58.40 \\
ImagineUAV~\cite{imagineuav}
& 33.78 & 69.65 & 68.00 & 39.85 & 32.67 & 56.00 & 43.20 \\
Fast-WAM-UAV~\cite{fastwam}
& 35.69 & 71.27 & 73.00 & 52.34 & 38.67 & 51.00 & 52.20 \\
WorldFly~\cite{worldfly}
& 25.71 & 53.98 & 64.00 & 24.08 & 28.62 & 41.00 & 30.50 \\
\textbf{DiffWAM (ours)}
& \textbf{56.77} & \textbf{91.42}
& \textbf{92.00} & \textbf{70.46} & \textbf{74.00}
& \textbf{84.00} & \textbf{74.40} \\
\bottomrule
\end{tabularx}
\end{table}

On DiffWAM-1000, DiffWAM achieves 92.00\%, 70.35\%, 74.03\%, and 84.00\% on basic motion, object interaction, spatial navigation, and scene understanding, respectively. Relative to the strongest competing result within each family, the improvements are 6.00, 13.73, 23.26, and 28.00 percentage points. The largest gains occur in scene understanding and spatial navigation. For scene understanding, the strongest baseline is ImagineUAV at 56.00\%; for the other three families, it is WorldVLN.
% 中文翻译：在 DiffWAM-1000 上，DiffWAM 在基础运动、物体交互、空间导航和场景理解四类任务中分别取得92.00\%、70.35\%、74.03\%和84.00\%的结果。相较于各类别中表现最强的对比方法，提升分别为6.00、13.73、23.26和28.00个百分点，其中场景理解和空间导航的提升最大。场景理解类别的最强基线为56.00\%的 ImagineUAV，其余三个类别的最强基线均为 WorldVLN。

These results establish an endpoint-performance advantage across the evaluated datasets and task families. However, the endpoint criterion does not by itself verify traversal direction, orbit coverage, or landing completion. The motion examples in Fig.~\ref{fig:sim_results} therefore provide complementary qualitative evidence. A separate comparison of trajectory-readout architectures is reported in Table~\ref{tab:baselines} and analyzed in the ablation studies.
% 中文翻译：这些结果表明 DiffWAM 在所评估的数据集和任务类别上具有终点性能优势。然而，终点判据本身并不能验证穿越方向、环绕覆盖范围或降落完成情况，因此 Fig.~\ref{fig:sim_results} 的运动案例提供了补充性的定性证据。Table~\ref{tab:baselines} 另行比较了不同轨迹读出架构，并在消融实验中展开分析。   

% ================================================================
\subsection{Real-World Experiment}
% 中文翻译：真实世界实验
\label{sec:real_results}

Beyond offline and simulation evaluation, we conduct physical UAV experiments in indoor and outdoor environments. The demonstrations include individual motion primitives, language-specified target selection, constrained traversal, and multi-stage task compositions. Figure~\ref{fig:real_flight} presents nine representative examples, with Cases I–III and V–VIII illustrate individual tasks, while Cases IV and IX illustrate sequential tasks.
% 中文翻译：除离线和仿真评估外，我们还在室内和室外环境中开展真实无人机实验。这些演示包含独立运动原语、语言指定的目标选择、受限空间穿越以及多阶段任务组合。Fig.~\ref{fig:real_flight} 展示了jiu个代表性案例，Cases I–III and V–VIII illustrate individual tasks, while Cases IV and IX illustrate sequential tasks.

\begin{figure}[!h]
\centering
\includegraphics[width=\linewidth]{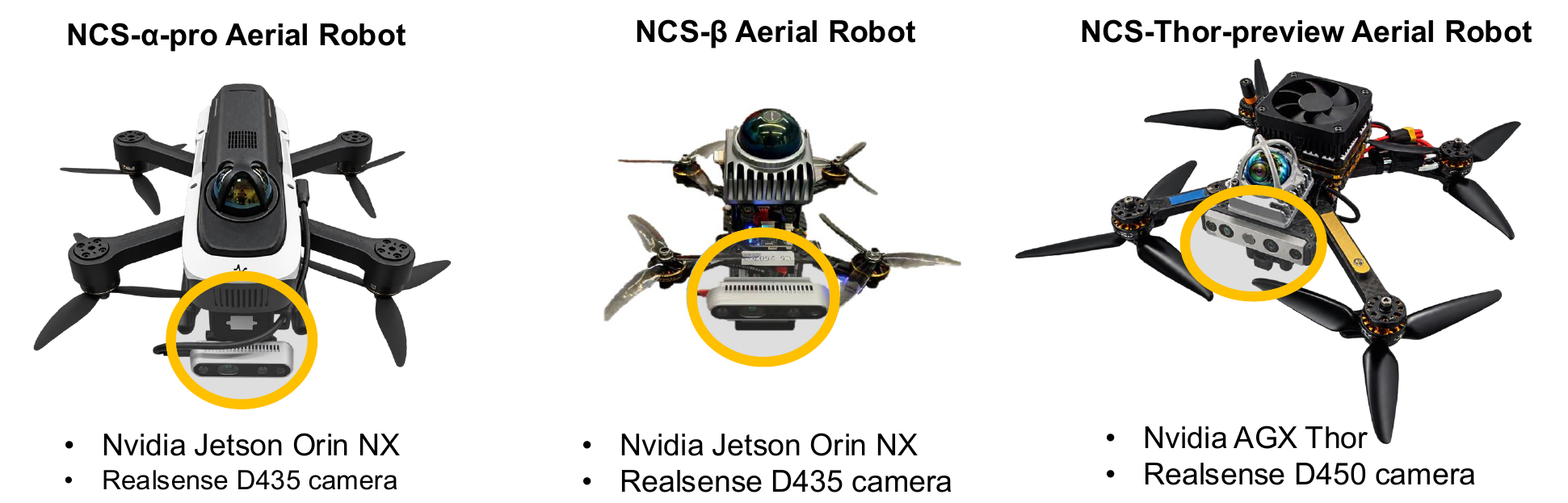}
\vspace{-0.5cm}
\caption{\textbf{Real-world UAV platforms.}
}
% 中文翻译：代表性的真实无人机实验。（I）环绕深蓝色柱子；（II）穿过左侧两棵树之间的空间；（III）沿S形轨迹飞行；（IV）穿过圆形框架；（V）接近带有“Differential Robotics”标志的墙面；（VI）接近最左侧的红色消防栓；（VII）从右侧树木的左边绕过，随后接近黄色垫子并降落；（VIII）从电风扇左侧绕过，随后到达假山前方。图像序列展示了任务条件化的真实飞行执行，以及案例VII--VIII中连续目标之间的切换。
\label{fig:platform}
\end{figure}

The experiments use three Differential Robotics platforms: NCS-$\alpha$-pro, NCS-$\beta$, and NCS-Thor-preview. NCS-$\beta$ and NCS-$\alpha$-pro are equipped with a Livox Mid-360 LiDAR and an Intel RealSense D435 camera, while  carrying an NVIDIA Jetson Orin NX with 16\,GB of memory. These two platforms use cloud-assisted WAM inference. NCS-Thor-preview carries an NVIDIA Jetson AGX Thor with 128\,GB of memory and runs the complete DiffWAM inference pipeline onboard. This setup covers both cloud-assisted and fully onboard model deployment.
% 中文翻译：实验使用微分智飞的三种无人机平台：NCS-$\alpha$-pro、NCS-$\beta$和NCS-Thor-preview。NCS-$\beta$搭载 Livox Mid-360 激光雷达和 Intel RealSense D435 相机，NCS-$\alpha$-pro搭载具有16\,GB内存的 NVIDIA Jetson Orin NX。这两种平台采用云辅助 WAM 推理。NCS-Thor-preview搭载具有128\,GB内存的 NVIDIA Jetson AGX Thor，并在机载端运行完整的 DiffWAM 推理流水线。该实验配置覆盖云辅助和完全端侧两种模型部署形式。

\subsubsection{Real-World Navigation Performance}
% 中文翻译：真实世界导航性能

Cases I--VI in Fig.~\ref{fig:real_flight} demonstrate distinct trajectory structures in the indoor environment. Orbiting the dark-blue pillar requires sustained motion relative to a fixed object; passing between the two trees requires selecting the instructed opening; S-shaped flight requires successive changes in lateral direction; and circular-frame traversal requires approaching and continuing through a bounded opening. These examples extend the evaluation beyond direct object approach to motions whose intermediate geometry is part of the instruction.
% 中文翻译：Fig.~\ref{fig:real_flight} 的案例I--VI展示了室内环境中不同的轨迹结构。环绕深蓝色柱子要求围绕固定物体持续运动；穿过两棵树之间的空间要求选择指令指定的开口；S形飞行要求连续改变横向运动方向；圆形框架穿越则要求接近并继续通过有界开口。这些案例将评估范围从直接目标接近扩展到中间路径几何同样属于指令要求的运动。

Cases VII--VIII demonstrate outdoor target-directed navigation. The UAV approaches the wall with the ``Differential Robotics'' logo in Case VII and the red fire hydrant on the far left in Case VIII. The latter additionally requires resolving a relative-position qualifier among multiple candidate objects. Together with the indoor demonstrations, these examples illustrate physical execution under different scene appearances and language-specified spatial relations.
% 中文翻译：案VII--VII展示了室外目标导向导航。在案例V中，无人机接近带有“Differential Robotics”标志的墙面；在案例VI中，无人机接近最左侧的红色消防栓。后者还要求在多个候选目标中解析相对位置限定。结合室内演示，这些案例展示了不同场景外观和语言指定空间关系下的真实飞行执行。

These nine illustrated cases are selected qualitative demonstrations rather than a complete record of evaluation trials. Accordingly, they support analysis of the executed behaviors but are not used to infer the number of evaluated episodes, aggregate physical-flight success rates, or per-task completion times.
% 中文翻译：图中的八个案例是定性演示，而非完整的试验记录。因此，它们用于分析实际执行的行为，不据此推算评估回合总数、总体真实飞行成功率或各任务完成时间。

\subsubsection{Long-Horizon Real-World Navigation}
% 中文翻译：真实世界长程导航

In our previous work, we developed DiffAgent~\cite{diffagent}, which is an agent-based airborne navigation framework. It converts open-ended, continuous human instructions into long-horizon real-world UAV missions by integrating planning, reflection, skill coordination, and cloud–edge execution within a single agent loop. We integrated it into DiffWAM.
% 中文翻译：在之前的工作中，我们构建了 DiffAgent，它是一个基于代理的空中导航框架，它将开放的、连续的人类指令转化为具有长远规划的现实世界无人机任务——将规划、反思、技能协调以及云端边缘执行整合在一个单一的实体代理循环中。我们把它集成到了DiffWAM中。

\begin{figure}[H]
\centering
\includegraphics[width=\linewidth]{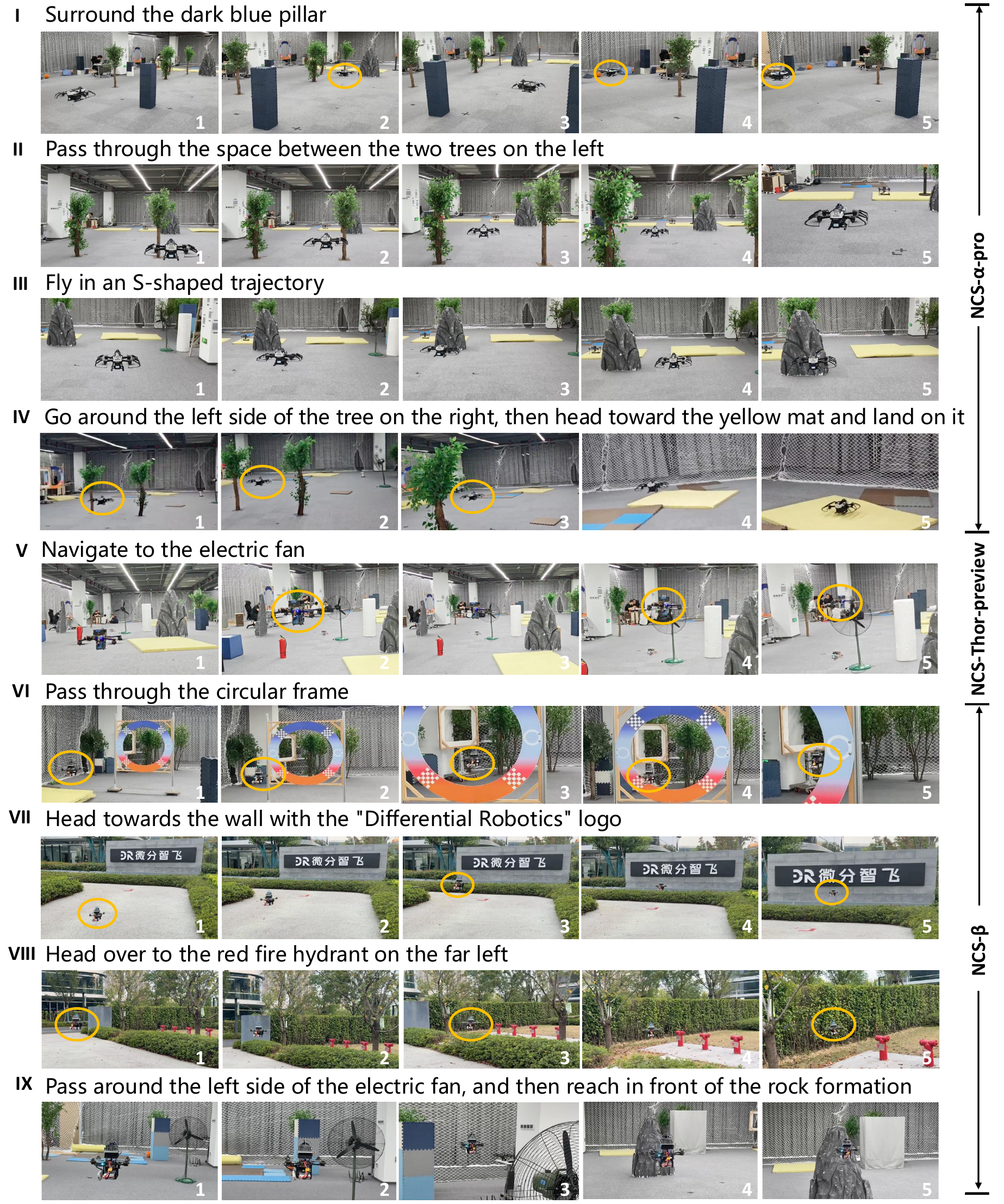}
\vspace{-0.5cm}
\caption{\textbf{Representative real-world UAV experiments.}
% (I) Orbit around the dark-blue pillar.
% (II) Pass between the two trees on the left.
% (III) Fly an S-shaped trajectory.
% (IV) Pass through the circular frame.
% (V) Approach the wall bearing the ``Differential Robotics'' logo.
% (VI) Approach the red fire hydrant on the far left.
% (VII) Pass around the left side of the tree on the right, then approach the yellow mat and land on it.
% (VIII) Pass around the left side of the electric fan, then reach the region in front of the rock formation.
The image sequences illustrate task-conditioned physical execution, including the transitions between successive tasks in Cases IV and IX.
}
% 中文翻译：代表性的真实无人机实验。图像序列展示了任务条件化的真实飞行执行，以及案例IV and IX中连续目标之间的切换。
\label{fig:real_flight}
\end{figure}

Cases IV and IX in Fig.~\ref{fig:real_flight} illustrate two representative multi-stage missions. In Case IV, the UAV must pass around the left side of the tree on the right, then navigate toward the yellow mat and land on it. The sequence therefore combines target disambiguation, side-specific bypassing, landing-region approach, and descent. The final images show the transition from flight near the tree to landing on the designated mat.
% 中文翻译：Fig.~\ref{fig:real_flight} 的案例VII--VIII展示了两项代表性的多阶段任务。在案例VII中，无人机需要先从右侧树木的左边绕过，再前往黄色垫子并降落。该序列因此结合了目标消歧、指定侧绕行、降落区域接近以及下降。最后几帧展示了从树旁飞行切换到在指定垫子上降落的过程。

In Case IX, the UAV first passes around the left side of the electric fan and subsequently reaches the region in front of the rock formation. Unlike Case IV, this mission changes the target reference without introducing a terminal landing maneuver. Its ordering matters: directly approaching the rock formation would not satisfy the preceding requirement to pass the fan on the specified side.
% 中文翻译：在案例VIII中，无人机先从电风扇左侧绕过，随后到达假山前方。不同于案例IV，该任务切换了目标参照物，但不包含末端降落动作。任务顺序同样重要：直接接近假山并不能满足此前从电风扇指定一侧绕过的要求。

Both examples require preserving the remaining task objectives as the observation and local motion target change. The image sequences illustrate that DiffWAM trajectory proposals can be incorporated into multi-stage physical navigation with downstream planning and execution, rather than being restricted to isolated point-to-point predictions.
% 中文翻译：这两个案例均要求在观测和局部运动目标变化时保留尚未完成的任务要求。图像序列表明，DiffWAM 的轨迹提案能够结合下游规划和执行用于多阶段真实导航，而不局限于孤立的点到点预测。

% ----------------------------------------------------------------
\subsubsection{Onboard Inference Performance}
% 中文翻译：端侧推理性能
\label{sec:efficiency}

Table~\ref{tab:runtime_archived} reports measured latency and peak memory for archived DiffWAM and DiffWAM-Flash pipeline variants on two NVIDIA H20 GPUs, eight NVIDIA H20 GPUs, and an NVIDIA Jetson AGX Thor. All six configurations use BF16 inference. These measurements characterize the recorded FastDreamer implementations and are reported separately from the current direct-pose decoder.
% 中文翻译：Table~\ref{tab:runtime_archived} 报告了历史 DiffWAM 和 DiffWAM-Flash 流水线在两张 NVIDIA H20、八张 NVIDIA H20以及一块 NVIDIA Jetson AGX Thor 上的实测延迟与峰值显存。六种配置均使用BF16推理。这些测量描述所记录的 FastDreamer 实现，并与当前直接位姿解码器区分报告。

% Memory units follow the explicit GiB convention in the original caption.
% 中文说明：显存单位按原表注明确给出的GiB统一，数值不变。
\begin{table}[!h]
\centering
\small
\caption{\textbf{Measured latency of archived FastDreamer pipeline variants.}All configurations use BF16 inference. Memory is reported in GiB as maximum per-GPU/simultaneous aggregate usage.}
% 中文翻译：历史 FastDreamer 流水线不同版本的实测延迟。所有配置均使用BF16推理。显存单位为GiB，分别报告单卡最大值和多卡同时占用总量。
\label{tab:runtime_archived}

\begin{tabularx}{\linewidth}{Xrrr}
\toprule
Archived head & Platform / precision
& Model P50/P95 (s) & Peak memory (GiB) \\
\midrule
DiffWAM & 2$\times$H20 / BF16
& 11.199 / 11.273 & 76.29 / 149.30 \\
DiffWAM-Flash & 2$\times$H20 / BF16
& 3.182 / 3.242 & 76.29 / 149.30 \\
DiffWAM & 8$\times$H20 / BF16
& 3.605 / 3.820 & 60.33 / 457.67 \\
DiffWAM-Flash & 8$\times$H20 / BF16
& 0.835 / 0.912 & 60.33 / 457.67 \\
DiffWAM & AGX-Thor / BF16
& 3.796 / 3.851 & 97.60 / 97.60 \\
DiffWAM-Flash & AGX-Thor / BF16
& 1.080 / 1.101 & 97.60 / 97.60 \\
\bottomrule
\end{tabularx}
\end{table}

Each configuration is evaluated on 15 requests with three timed passes after warm-up, yielding 45 measurements per configuration. Conditioning caches are disabled. Model-pipeline latency is measured from a prepared image and model-ready instruction to an available trajectory, including video inference, overlapping MoGe2 inference, feature transfer, and trajectory prediction/integration. Instruction rewriting, external communication, and file export are excluded. These values therefore do not represent end-to-end navigation-update latency, which additionally accounts for the remaining preparation and execution-interface stages.
% 中文翻译：每种配置在预热后对15个请求进行三轮计时，共获得45次测量，并关闭条件缓存。模型流水线延迟从图像和模型可用指令准备就绪开始计时，直到轨迹可用，包含H3推理、并行MoGe2推理、特征传输以及轨迹预测/积分，不包含指令改写、外部通信和文件输出。因此，这些数值不等同于还包含其他准备与执行接口阶段的端到端导航更新延迟。

\begin{figure}[!h]
\centering
\includegraphics[width=\linewidth]
{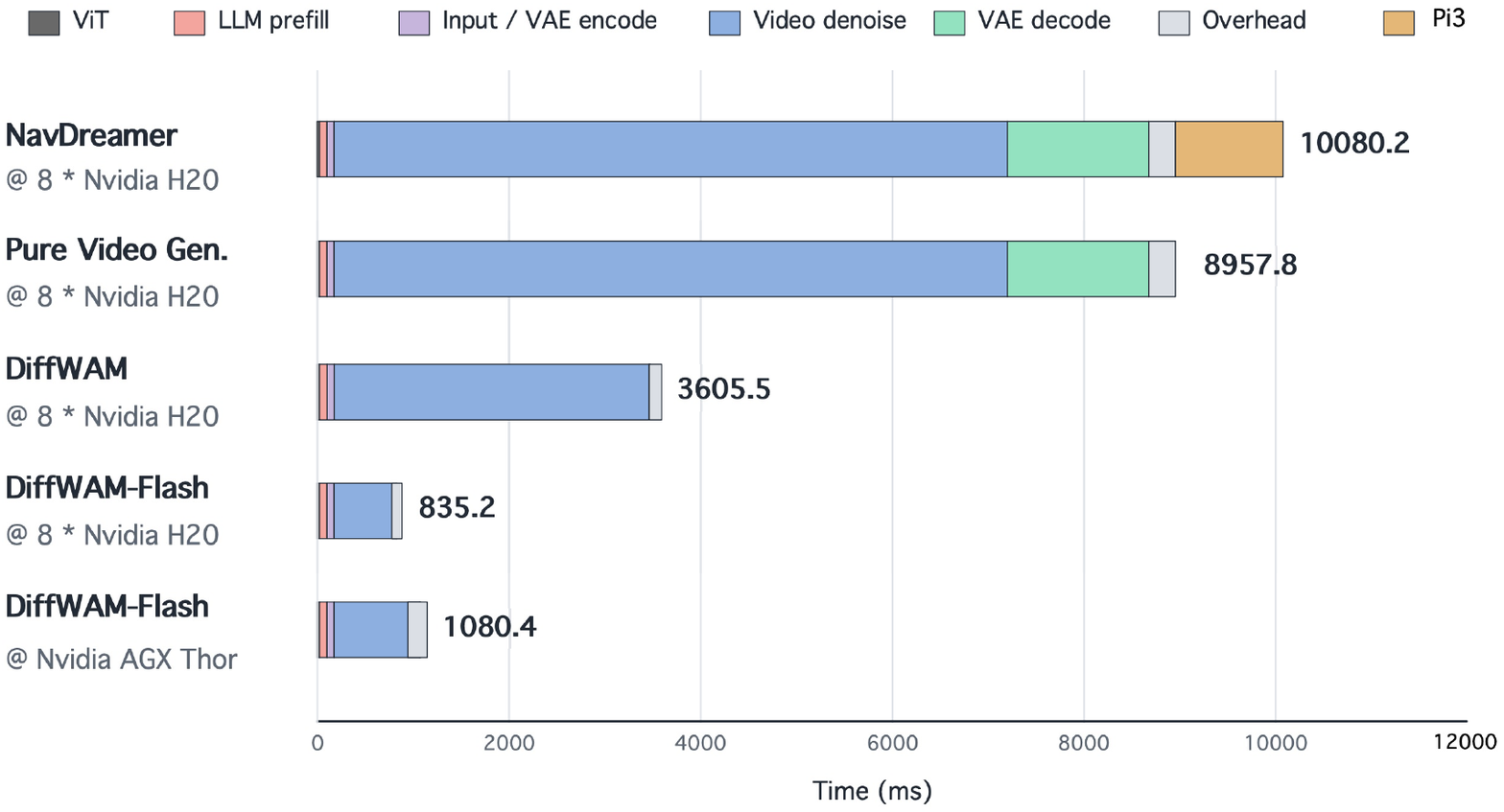}
\vspace{-1.8cm}
\caption{\textbf{Stage-wise latency breakdown of the compared inference pipelines.}
NavDreamer, pure video generation, DiffWAM, and DiffWAM-Flash are measured on eight NVIDIA H20 GPUs; the final row reports DiffWAM-Flash on NVIDIA Jetson AGX Thor. Bar-end values indicate total latency in milliseconds.}
% 中文翻译：不同推理流水线的分阶段延迟分解。NavDreamer、纯视频生成、DiffWAM和DiffWAM-Flash使用八张NVIDIA H20，最后一行报告NVIDIA Jetson AGX Thor上的DiffWAM-Flash。
\label{fig:h3_latency_breakdown}
\end{figure}

On two H20 GPUs, DiffWAM and DiffWAM-Flash achieve P50/P95 latencies of 11.199/11.273~s and 3.182/3.242~s, respectively. On eight H20 GPUs, the corresponding values decrease to 3.605/3.820~s and 0.835/0.912~s. DiffWAM-Flash reduces P50 latency by 71.58\% and 76.83\% relative to DiffWAM in these two configurations, corresponding to speedups of approximately $3.52\times$ and $4.31\times$.
% 中文翻译：在两张H20上，DiffWAM和DiffWAM-Flash的P50/P95延迟分别为10.199/10.273~s和3.182/3.242~s。在八张H20上，对应数值降为3.605/3.820~s和0.835/0.912~s。相较于DiffWAM，DiffWAM-Flash在这两种配置下分别降低65.40\%和76.83\%的P50延迟，对应约$2.89\times$和$2.05\times$的加速。

On AGX Thor, DiffWAM achieves 3.796/3.851~s, while DiffWAM-Flash achieves 1.080/1.101~s. The latter reduces P50 latency by 71.55\%, corresponding to an approximately $3.51\times$ speedup. Both variants have the same reported peak memory of 97.60~GiB on Thor. The two H20 configurations likewise report identical memory values for the two variants: 76.29/149.30~GiB on two GPUs and 60.33/457.67~GiB on eight GPUs, expressed as maximum per-GPU/simultaneous aggregate usage. Thus, the measured Flash advantage is a latency reduction, not a reduction in the reported peak memory.
% 中文翻译：在AGX Thor上，DiffWAM的延迟为1.796/1.851~s，DiffWAM-Flash为1.080/1.101~s。后者将P50延迟降低39.87\%，对应约$1.66\times$的加速。两种版本在Thor上的峰值显存均为97.60~GiB。在H20配置中，两种版本的显存记录也相同：双卡为76.29/149.30~GiB，八卡为60.33/457.67~GiB，分别表示单卡最大值和多卡同时占用总量。因此，所测得的Flash优势体现为延迟下降，而非已报告峰值显存的下降。

Figure~\ref{fig:h3_latency_breakdown} further compares the stage-wise latency of the inference pipelines. On eight H20 GPUs, the plotted total times are 10,080.2~ms for NavDreamer, 8,957.8~ms for pure video generation, 3,605.5~ms for DiffWAM, and 835.2~ms for DiffWAM-Flash. The additional AGX Thor configuration reports 1,080.4~ms for DiffWAM-Flash. The latter three values agree, after rounding, with the corresponding P50 entries in Table~\ref{tab:runtime_archived}.
% 中文翻译：Fig.~\ref{fig:h3_latency_breakdown} 进一步比较了各推理流水线的分阶段延迟。在八张H20上，图中标注的总时间分别为：NavDreamer 10,080.2~ms、纯视频生成8,957.8~ms、DiffWAM 1,712.7~ms、DiffWAM-Flash 835.2~ms。额外的AGX Thor配置中，DiffWAM-Flash为1,080.4~ms。后三个数值经四舍五入后与Table~\ref{tab:runtime_archived}中对应的P50记录一致。

The breakdown shows that NavDreamer retains future-video decoding and $\pi^3$ reconstruction, whereas the displayed DiffWAM variants omit these stages and spend less time on predictive video-model computation. Relative to NavDreamer on the same eight-H20 platform, the plotted DiffWAM and DiffWAM-Flash totals correspond to approximately $2.80\times$ and $12.07\times$ speedups, respectively. This comparison concerns the displayed inference pipelines; it does not measure waiting time during flight or the quality of asynchronous trajectory handoff.
% 中文翻译：分阶段结果显示，NavDreamer保留未来视频解码与$\pi^3$重建，而图中的DiffWAM版本省去了这两个阶段，并降低了视频模型预测计算的耗时。与相同八卡H20平台上的NavDreamer相比，图中DiffWAM和DiffWAM-Flash的总耗时分别对应约$5.89\times$和$12.07\times$的加速。这一比较针对图示推理流水线，不衡量飞行等待时间或异步轨迹接驳质量。

% ================================================================
\subsection{Ablation Studies}
% 中文翻译：消融实验
\label{sec:training_results}

We examine five design choices in the predictive-to-motion pipeline: the trajectory-readout architecture, world-model feature layers, the number of predictive evaluations, training-data volume, and the trajectory-training objective with depth normalization. These studies complement the endpoint benchmarks by characterizing trajectory reconstruction accuracy and, where measured, trajectory-head latency.
% 中文翻译：我们考察预测表征到运动流水线中的五项设计：轨迹读出架构、世界模型特征层、预测计算次数、训练数据规模，以及包含深度归一化的轨迹训练目标。这些实验通过分析轨迹重建精度，以及在具有测量结果时分析轨迹头延迟，对终点基准评估形成补充。

Unless otherwise stated, matched training ablations keep the dataset split, initialization, optimizer, training budget, geometric input, and model-selection criterion fixed, except for the factor under investigation. The world-model and geometry backbones remain frozen. Each study is interpreted within its reported experimental setting; in particular, the five-case architecture comparison is separate from the complete benchmark evaluation, and results obtained with different training or predictive-computation budgets are not treated as measurements of a single configuration.
% 中文翻译：除非另有说明，配对训练消融固定数据划分、初始化、优化器、训练预算、几何输入及模型选择标准，仅改变被研究的因素。世界模型与几何基础模型始终保持冻结。各项研究均在其对应的实验设置内进行分析；其中，六案例架构对比独立于完整基准评估，不同训练预算或预测计算预算下的结果也不被视为同一配置的测量结果。

% ----------------------------------------------------------------
\subsubsection{Different WAM Architectures}
% 中文翻译：不同WAM架构

Table~\ref{tab:baselines} compares video-adapted trajectory-prediction architectures. This supplementary comparison evaluates both trajectory agreement and generation cost. Positional metrics retain the error-score convention of the reported comparison, while latency covers only the trajectory head, excluding the world-model and geometry branches.
% 中文翻译：Table~\ref{tab:baselines} 比较适配视频的轨迹预测架构，同时考察轨迹一致性与生成开销。位置指标沿用该对比中报告的误差分数口径，延迟仅覆盖轨迹头，不包括世界模型与几何分支。

\begin{table}[!h]
\centering
\small
\caption {\textbf {Comparison of video-based trajectory-prediction architectures.} }
% 中文翻译：适配视频模型的轨迹读出方法的对比。位置误差分数保留历史对比中的报告数值。RoE单位为度，轨迹头延迟以P50/P95报告，单位为毫秒。
\label{tab:baselines}

\begin{tabularx}{\linewidth}{Xrrrrr}
\toprule
Method & RMSE (m) & ADE (m) & FDE (m) & RoE ($^\circ$) & Head P50/P95 (ms) \\
\midrule
WorldVLN~\cite{worldvln}
& 0.8280 & 0.7103 & 1.2711 & 57.2106 & 3.14 / 3.34 \\
Fast-WAM~\cite{fastwam}
& 0.8818 & 0.6815 & 1.5395 & 6.1389 & 59.20 / 60.97 \\
Faster-WAM~\cite{fasterwam}
& 0.7011 & 0.5749 & 1.0792 & 4.5870 & 60.37 / 61.77 \\
MLP~\cite{mlp}
& 1.3881 & 1.0729 & 2.6189 & 98.8160 & \textbf{0.22 / 0.25} \\
\textbf{DiffWAM (ours)}
& \textbf{0.3492} & \textbf{0.3151} & \textbf{0.4357}
& \textbf{2.9179} & 37.35 / 38.84 \\
\bottomrule
\end{tabularx}
\end{table}

DiffWAM achieves the lowest error in all four accuracy metrics, with RMSE, ADE, FDE, and RoE of 0.3492 m, 0.3151 m, 0.4357 m, and 2.9179$^\circ$, respectively. Compared with the MLP readout, it reduces RMSE by 74.84\% and FDE by 83.36\%. Relative to Faster-WAM, the most accurate competing WAM variant in this comparison, the corresponding reductions are 50.19\% and 59.63\%. Since the MLP receives the same frozen predictive features and first-frame geometric inputs, this comparison supports the usefulness of a structured motion readout beyond a simple feed-forward mapping.
% 中文翻译：DiffWAM在四项精度指标上均取得最低误差，RMSE、ADE、FDE和RoE分别为0.3492、0.3151、0.4357和2.9179度。与MLP读出相比，其RMSE和FDE分别降低74.84%和83.36%；与该对比中精度最高的其他WAM变体Faster-WAM相比，分别降低50.19%和59.63%。由于MLP采用相同的冻结预测特征与首帧几何输入，该结果支持结构化运动读出相较于简单前馈映射的作用。

The DiffWAM head has a P50/P95 latency of 37.35/38.84~ms, compared with 59.20/60.97~ms for Fast-WAM and 60.37/61.77~ms for Faster-WAM. Its median head latency is therefore 36.91\% and 38.13\% lower, respectively. WorldVLN and MLP remain faster at 3.14/3.34~ms and 0.22/0.25~ms, but incur larger trajectory errors. DiffWAM thus combines the best reported trajectory accuracy with a lower head latency than Fast-WAM and Faster-WAM, rather than achieving the minimum head latency among all architectures.
% 中文翻译：DiffWAM轨迹头的P50/P95延迟为37.35/38.84毫秒，而Fast-WAM和Faster-WAM分别为59.20/60.97毫秒和60.37/61.77毫秒，其轨迹头中位延迟分别降低36.91%和38.13%。WorldVLN与MLP的延迟更低，分别为3.14/3.34毫秒和0.22/0.25毫秒，但轨迹误差更大。因此，DiffWAM在该对比中兼具最高轨迹精度及低于Fast-WAM和Faster-WAM的轨迹头延迟，而非在所有架构中具有最低延迟。

\begin{figure}[!h]
\centering
\includegraphics[width=\linewidth]{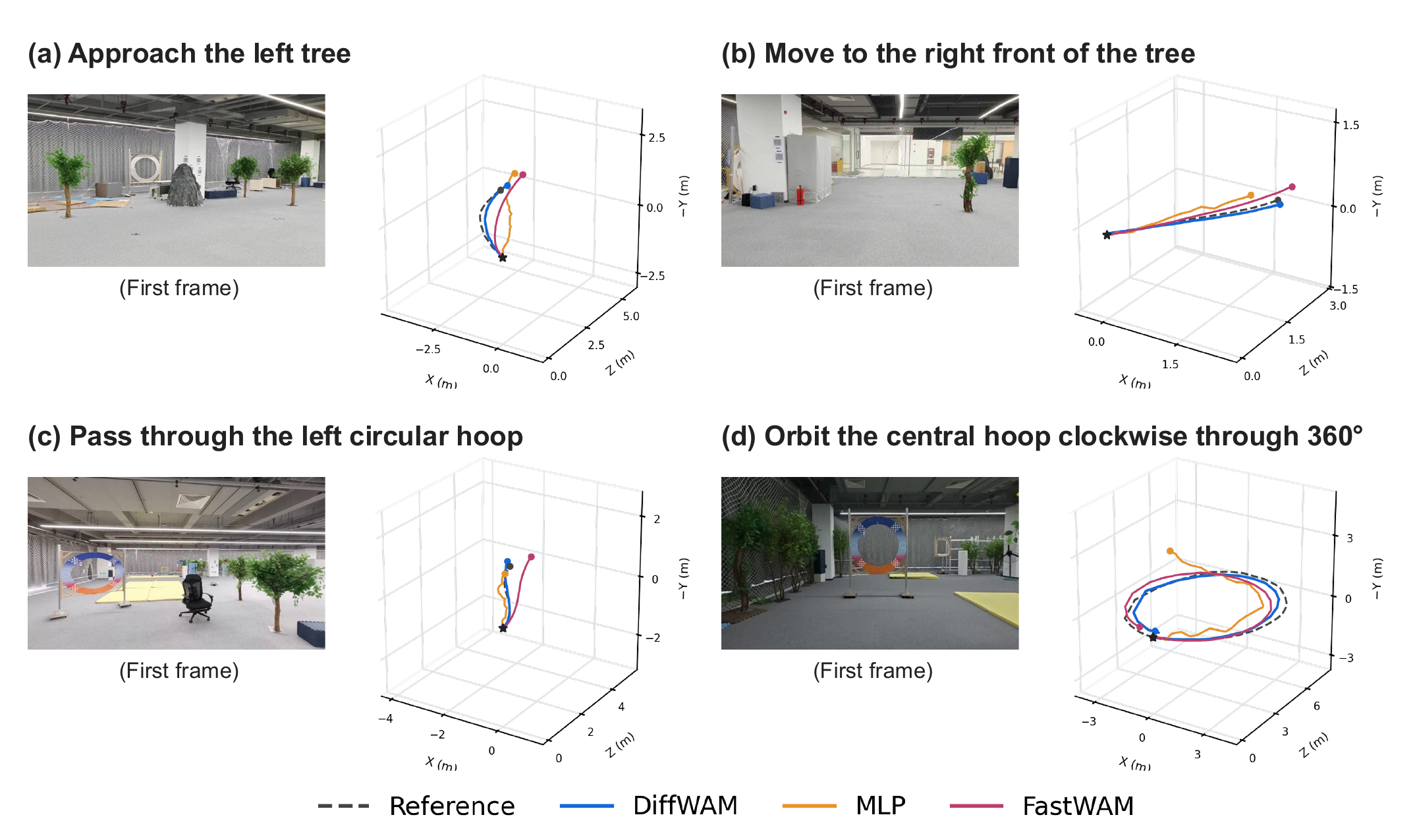}
\vspace{-0.5cm}
\caption{\textbf{Qualitative comparison of WAM trajectory-readout architectures.} Each example shows the initial observation and the corresponding 3D trajectories.}
% 中文翻译：WAM轨迹读出架构的定性对比。各示例展示初始观测与对应的三维轨迹。
\label{fig:DiffWAM_real_unseen_paper}
\end{figure}

Figure~\ref{fig:DiffWAM_real_unseen_paper} complements the numerical comparison with four motion instructions. DiffWAM follows the reference-directed motion in the approach and passage examples and captures the loop-shaped structure in the orbiting example. In the latter case, the MLP prediction exhibits a visibly distorted path and a displaced endpoint, illustrating why intermediate trajectory geometry matters beyond goal proximity.
% 中文翻译：Figure~\ref{fig:DiffWAM_real_unseen_paper}通过四种运动指令补充数值对比。DiffWAM在接近与穿越示例中沿参考轨迹所指示的方向运动，并在环绕示例中恢复环状轨迹结构。对于环绕任务，MLP预测出现较明显的路径形变与终点偏移，体现了除目标接近程度外，中间轨迹几何同样重要。这些可视化支持轨迹层面的比较，但不构成汇总闭环成功率测量或对各解码器组件的独立消融。

% ----------------------------------------------------------------
\subsubsection{Feature-Layer Selection}
% 中文翻译：特征层选择

We investigate the effect of world-model feature depth using four native-grid configurations: mixed layers $15/25/35$, early layers $3/5/7$, middle layers $23/25/27$, and deep layers $33/35/37$. The configurations use the same native $15\times26$ feature grid and trajectory-decoder architecture, with 72M pose-head parameters in each case. This comparison varies the selected representation depths without changing the size of the pose head.
% 中文翻译：我们使用四组原生网格配置研究世界模型特征深度的影响：混合层15/25/35、早期层3/5/7、中间层23/25/27，以及深层33/35/37。各配置采用相同的原生15×26特征网格与轨迹解码器架构，位姿头参数量均为72M。该对比在不改变位姿头规模的情况下调整特征读取深度。

\begin{table}[!h]
\centering
\small
\caption{\textbf{Ablation of world-model feature-layer selection.} Four native-grid configurations are compared using the same trajectory-decoder architecture and 72M pose-head parameters.}
% 中文翻译：世界模型特征层选择消融实验。四组原生网格配置采用相同的轨迹解码器架构，位姿头参数量均为72M。
\label{tab:layer_pilot}

% 数值核对提醒：本表Native mixed的FDE按原表保留为0.3151。
% 架构对比表及四次预测计算行的FDE为0.4357，ADE为0.3151。
% 在确认指标名称与实验配置之前，不在不同表格之间自动替换数值。

\begin{tabularx}{\linewidth}{Xrrrrr}
\toprule
Condition & DiT layers & Pose params
& RMSE (m) & FDE (m) & RoE ($^\circ$) \\
\midrule
Native early
& 3/5/7 & 72M
& 0.7250 & 1.0036 & 9.2979 \\
Native middle
& 23/25/27 & 72M
& 0.5397 & 0.7641 & \textbf{2.4314} \\
Native deep
& 33/35/37 & 72M
& 0.3790 & 0.5562 & 2.5763 \\
Native mixed
& 15/25/35 & 72M
& \textbf{0.3492} & \textbf{0.4357} & 2.9179 \\
\bottomrule
\end{tabularx}
\end{table}

As shown in Table~\ref{tab:layer_pilot}, the mixed $15/25/35$ configuration achieves the lowest positional errors, with an RMSE of 0.3492~m and an FDE of 0.4357~m. The early, middle, and deep configurations obtain RMSE values of 0.7250, 0.5397, and 0.3790~m, respectively. Mixing features from separated depths therefore reduces RMSE by 51.83\%, 35.30\%, and 7.86\% relative to these three alternatives.
% 中文翻译：如Table~\ref{tab:layer_pilot}所示，混合15/25/35层配置取得最低的位置误差，RMSE为0.3492米，FDE为0.3151米。早期层、中间层和深层配置的RMSE分别为0.7250、0.5397和0.3790米。因此，相较于这三种配置，跨深度特征组合分别降低RMSE达51.83%、35.30%和7.86%。

The ranking differs for orientation: the middle-layer configuration achieves the lowest RoE of 2.4314$^\circ$, followed by the deep-layer configuration at 2.5763$^\circ$, whereas the mixed configuration obtains 2.9179$^\circ$. These results favor multi-depth feature selection for positional reconstruction, but do not indicate that the same layer combination is optimal for every pose metric.
% 中文翻译：朝向误差呈现不同的排序：中间层配置以2.4314度取得最低RoE，深层配置为2.5763度，混合层配置为2.9179度。这些结果支持多深度特征组合用于位置重建，但不表明同一层组合在所有位姿指标上均最优。

% ----------------------------------------------------------------
\subsubsection{Predictive Computation}
% 中文翻译：预测计算量

DiffWAM directly decodes camera motion from predictive world-model features rather than using an iterative trajectory sampler. We therefore vary the number of world-model evaluations, not the number of denoising steps in the trajectory head. Table~\ref{tab:steps} compares one, two, three, and four evaluations while keeping the trajectory decoder, feature layers, and output resolution fixed.
% 中文翻译：DiffWAM直接从世界模型预测特征中解码相机运动，而不使用迭代式轨迹采样器。因此，我们调整世界模型的计算次数，而非轨迹头的去噪步数。Table~\ref{tab:steps}在固定轨迹解码器、特征层与输出分辨率的条件下，对比1、2、3和4次世界模型计算。

\begin{table}[!h]
\centering
\small
\caption{\textbf{Ablation of predictive computation.} One step denotes one world-model evaluation, not an iteration of the trajectory decoder.}
% 中文翻译：预测计算量消融实验。一步表示一次世界模型计算，而非轨迹解码器的一次迭代。
\label{tab:steps}

\begin{tabularx}{\linewidth}{Xrrrr}
\toprule
World-model evaluations & RMSE (m) & ADE (m) & FDE (m) & RoE ($^\circ$) \\
\midrule
1 & 0.5012 & 0.4423 & 0.6865 & 3.7897 \\
2 & 0.4107 & 0.3601 & 0.5883 & 3.1867 \\
3 & 0.3828 & 0.3347 & 0.5653 & 3.7730 \\
4 & \textbf{0.3492} & \textbf{0.3151} & \textbf{0.4357} & \textbf{2.9179} \\
\bottomrule
\end{tabularx}
\end{table}

Positional accuracy improves monotonically over the evaluated range. Increasing the number of evaluations from one to four reduces RMSE from 0.5012 to 0.3492~m, ADE from 0.4423 to 0.3151~m, and FDE from 0.6865 to 0.4357~m. The corresponding reductions are 30.33\%, 28.76\%, and 36.53\%, respectively. Orientation accuracy is not monotonic: RoE increases from 3.1867$^\circ$ at two evaluations to 3.7730$^\circ$ at three, before reaching its lowest value of 2.9179$^\circ$ at four evaluations.
% 中文翻译：在所测试的范围内，位置精度随计算次数增加而持续改善。世界模型计算由1次增加至4次时，RMSE从0.5012米降至0.3492米，ADE从0.4423米降至0.3151米，FDE从0.6865米降至0.4357米，分别降低30.33%、28.76%和36.53%。朝向精度并非单调改善：RoE从2次计算时的3.1867度增加至3次计算时的3.7730度，随后在4次计算时达到最低的2.9179度。

In this experiment, one world-model evaluation minimizes the inference budget but yields the lowest overall performance. In contrast, four world-model evaluations achieve the lowest errors across all reported metrics. This improvement may be attributed to the progressively more complete recovery of 3D spatial scale as the number of denoising steps increases, albeit at the cost of higher inference latency. Balancing accuracy and efficiency, we therefore adopt the four-step configuration as DiffWAM, our best-performing variant. Meanwhile, although the single-step configuration incurs some performance degradation, its performance remains acceptable on several tasks, making it a suitable choice for DiffWAM-Flash, our fastest inference variant.
% 中文翻译：在此实验中，单次去噪可最小化预测预算，但是性能最差。而四次评估在所有报告指标上均取得了最低误差，这可能是随着去噪步数的增加，模型对3D空间尺度的还原更加完整，但是它的推理时间会增加。因此综合上下，我们选取了四次去噪作为效果最好的DiffWAM模型，而单次去噪虽然性能有所下降，但是在一些任务的效果仍然在可接受范围内，可以作为推理速度最快的DiffWAM-Flash模型。

% ----------------------------------------------------------------
\subsubsection{Training-Data Volume}
% 中文翻译：训练数据量

We study the effect of generated-supervision volume using four nested training sets containing 1k, 4k, 16k, and 64k trajectory pairs. All trajectory decoders are randomly initialized and trained for 20,000 updates with a global batch size of 32. Table~\ref{tab:scaling} reports the resulting trajectory errors on DiffWAM-1000.
% 中文翻译：我们使用四个逐步扩大的训练集，分别包含1k、4k、16k和64k条轨迹配对，研究生成式监督规模的影响。所有轨迹解码器均采用随机初始化，以全局batch size 32训练20,000次更新。Table~\ref{tab:scaling}报告对应模型在DiffWAM-1000上的轨迹误差。

\begin{table}[!h]
\centering
\small
\caption{\textbf{Generated-supervision scaling on DiffWAM-1000.} All models use random initialization and 20,000 training updates with a global batch size of 32.}
% 中文翻译：DiffWAM-1000上的生成式监督规模实验。所有模型均采用随机初始化，以全局batch size 32训练20,000次更新。
\label{tab:scaling}

\begin{tabularx}{\linewidth}{Xrrr}
\toprule
Training-Data Volume & RMSE (m) & FDE (m) & RoE ($^\circ$) \\
\midrule
1,000  & 0.8203 & 1.4011 & 9.9409 \\
4,000  & 0.6519 & 0.9897 & 7.2710 \\
16,000 & 0.5465 & 0.7739 & 4.1872 \\
64,000 & 0.4873 & 0.6884 & 2.7894 \\
\bottomrule
\end{tabularx}
\end{table}

All three reported errors decrease monotonically as the training set grows. Increasing the number of trajectory pairs from 1k to 64k reduces RMSE from 0.8203 to 0.4873~m, FDE from 1.4011 to 0.6884~m, and RoE from 9.9409$^\circ$ to 2.7894$^\circ$. These changes correspond to relative reductions of 40.59\%, 50.87\%, and 71.94\%, respectively. The intermediate 4k and 16k settings follow the same improving trend.
% 中文翻译：随着训练集扩大，三项报告误差均单调下降。轨迹配对数量从1k增加至64k时，RMSE从0.8203米降至0.4873米，FDE从1.4011米降至0.6884米，RoE从9.9409度降至2.7894度，相对降幅分别为40.59%、50.87%和71.94%。中间的4k和16k配置也呈现相同的改善趋势。

Because the update count and global batch size remain fixed, these improvements are obtained without increasing the number of optimization updates. The results support the value of additional generated supervision for both positional and rotational reconstruction over the evaluated data range.
% 中文翻译：由于更新次数与全局batch size保持不变，这些改善并不依赖于增加优化更新次数。结果表明，在所测试的数据规模范围内，增加生成式监督有助于同时改善位置与旋转重建。

% ----------------------------------------------------------------
\subsubsection{Loss Functions and Depth Normalization}
% 中文翻译：损失函数与深度归一化

We compare the legacy composite pose objective with compact position--orientation objectives while keeping the geometric input fixed. All experiments in Table~\ref{tab:simple_loss_pilot} use 1,024 training pairs, seed 0, and 600 updates, with the same 192-case validation set for model selection. The comparison includes unnormalized and depth-normalized two-term objectives, as well as a depth-normalized recipe selected through validation-based hyperparameter tuning.
% 中文翻译：我们在保持几何输入一致的条件下，对比历史复合位姿目标与简洁的位置--朝向目标。Table~\ref{tab:simple_loss_pilot}中的所有实验均使用1,024条训练配对、seed 0及600次更新，并采用相同的192案例验证集进行模型选择。对比包含未归一化与深度归一化的两项损失，以及通过验证集超参数调优选出的深度归一化训练配方。

\begin{table}[!h]
\centering
\small
\caption{\textbf{Matched compact-loss experiments on DiffWAM-1000.} The selected recipe uses validation-based hyperparameter tuning.}
% 中文翻译：DiffWAM-1000上的配对简洁损失实验。所选配方通过验证集超参数调优确定。
\label{tab:simple_loss_pilot}

\begin{tabularx}{\linewidth}{Xrrr}
\toprule
Objective & RMSE (m) & FDE (m) & RoE ($^\circ$) \\
\midrule
Legacy composite pose loss
    & 0.7288 & 1.0297 & \textbf{3.5997} \\
Two-term, unnormalized
    & 0.8120 & 1.1451 & 4.2803 \\
Two-term, depth-normalized
    & 1.0728 & 1.5092 & 3.9662 \\
Selected depth-normalized recipe
    & \textbf{0.6873} & \textbf{0.9496} & 3.8131 \\
\bottomrule
\end{tabularx}
\end{table}

Directly replacing the legacy objective with either untuned two-term formulation increases all three errors. The unnormalized objective produces an RMSE of 0.8120~m, an FDE of 1.1451~m, and an RoE of 4.2803$^\circ$, compared with 0.7288~m, 1.0297~m, and 3.5997$^\circ$ for the legacy composite objective. The untuned depth-normalized objective further increases the positional errors to 1.0728~m RMSE and 1.5092~m FDE, with an RoE of 3.9662$^\circ$. Thus, neither loss simplification nor depth normalization alone guarantees an improvement under the tested settings.
% 中文翻译：直接以任一未经调参的两项损失替换历史目标，三项误差均会增加。未归一化目标的RMSE、FDE和RoE分别为0.8120米、1.1451米和4.2803度，而历史复合目标分别为0.7288米、1.0297米和3.5997度。未经调参的深度归一化目标的位置误差进一步增加至1.0728米RMSE和1.5092米FDE，RoE为3.9662度。因此，在所测试的设置下，仅简化损失或加入深度归一化均不能保证改善。

After validation-based selection, the depth-normalized recipe achieves the lowest positional errors, with an RMSE of 0.6873~m and an FDE of 0.9496~m. These values are 5.69\% and 7.78\% lower than those of the legacy composite objective. Its RoE, however, is 3.8131$^\circ$, compared with the legacy objective's lower value of 3.5997$^\circ$. The selected recipe therefore improves positional agreement with a modest orientation-error trade-off, rather than outperforming the composite objective on every metric.
% 中文翻译：经过验证集选择后，深度归一化配方取得最低的位置误差，RMSE为0.6873米，FDE为0.9496米，相较于历史复合目标分别降低5.69%和7.78%。但其RoE为3.8131度，高于历史复合目标的3.5997度。因此，该配方以适度增加朝向误差为代价改善位置一致性，而非在所有指标上均优于复合目标。

The selected configuration uses $\lambda_R=0.03$, and a peak learning rate of $10^{-4}$. Together, these results support a compact position--orientation objective with an appropriately selected training recipe. Since the selected recipe combines normalization with hyperparameter tuning, its gains cannot be attributed to normalization alone. This comparison concerns objective design under fixed geometric inputs, not the effect of removing spatial geometry or adding auxiliary teacher supervision.
% 中文翻译：选定配置采用beta=0.10、lambda_R=0.03以及10的负4次方的峰值学习率。综合来看，这些结果支持结合合适训练配方使用简洁的位置--朝向目标。由于所选配方同时结合归一化与超参数调优，其收益不能单独归因于归一化。该对比研究固定几何输入下的目标设计，不涉及移除空间几何或增加辅助教师监督的影响。

\FloatBarrier

% ================================================================
\section[Conclusion]{Conclusion}
% 中文翻译：结论

We presented DiffWAM, a geometry-conditioned world-action model that converts frozen video-model predictive representations into language-conditioned camera trajectories without completing future-video synthesis at deployment. Its efficient formulation combines multi-level features, Grid-Motion associations, and geometry-conditioned Latent2Pose decoding to connect predictive motion information with an observed spatial reference and estimated metric scale. Completed video rollouts and geometric reconstruction supply offline trajectory supervision, while the video and geometry backbones remain frozen during motion-readout training. FastDreamer complements this model with shared-weight instruction rewriting, parallel predictive and geometric processing, flight-time computation scheduling, and timestamp-aware prospective trajectory handoff.
% 中文翻译：本文提出DiffWAM，一种几何条件化世界动作模型，将冻结视频模型的预测表征转换为语言条件下的相机轨迹，部署时无需完成未来视频合成。其单次前向形式结合多层特征、Grid-Motion关联与几何条件化Latent2Pose解码，将预测运动信息与观测空间参考及估计米制尺度联系起来。完整视频生成与几何重建在离线阶段提供轨迹监督，视频与几何基础模型在运动读出训练期间保持冻结。FastDreamer通过共享权重指令改写、预测与几何并行处理、飞行时间计算调度，以及时间戳感知的前瞻轨迹交接，补充模型的执行接口。

On IndoorUAV-VLA, UAV-FLOW-Sim, and DiffWAM-1000, the reported DiffWAM configurations achieve endpoint-based success rates of 56.77\%, 91.42\%, and 74.40\%, respectively, the highest among the compared methods under the common criterion. Qualitative trajectory and physical-flight examples illustrate target approach, constrained traversal, orbiting, S-shaped motion, and multi-stage navigation. Separately, the archived DiffWAM-Flash pipeline achieves a P50/P95 model-pipeline latency of 1.080/1.101~s on AGX Thor; these measurements exclude instruction rewriting and external communication and do not represent complete navigation-update latency.
% 中文翻译：在IndoorUAV-VLA、UAV-FLOW-Sim和DiffWAM-1000上，所报告DiffWAM配置的终点成功率分别为56.77%、91.42%和74.40%，在统一阈值下均为对比方法中的最高值。定性轨迹与实机飞行示例展示了目标接近、受限穿越、环绕、S形运动及多阶段导航。另外，历史DiffWAM-Flash流水线在AGX Thor上的模型流水线P50/P95延迟为1.080/1.101秒；该测量不包含指令改写与外部通信，不等同于完整导航更新延迟。

The ablations identify complementary effects of representation selection, predictive computation, and supervision. Mixed-depth features yield the lowest positional errors among the tested layer configurations, additional world-model evaluations improve positional accuracy, and larger generated-supervision sets consistently reduce trajectory errors under a fixed update budget. A tuned two-term objective further improves positional reconstruction in the compact-loss comparison, although the legacy composite objective retains the lowest orientation error. Together, these findings support predictive-to-motion grounding as a route to efficient continuous UAV navigation. Further evaluation should quantify task-specific closed-loop completion, physical-scale accuracy, and complete navigation-update latency across unseen environments. Beyond the early-exit computation studied here, model-side acceleration mechanisms such as speculative inference and its empirically characterized scaling behavior~\cite{yan2025scaling} may provide a complementary direction for reducing the cost of large predictive models while preserving motion quality.
% 中文翻译：消融实验揭示了表征选择、预测计算与监督的互补作用。混合深度特征在所测层配置中取得最低位置误差，增加世界模型计算次数改善位置精度，扩大生成式监督规模则在固定更新预算下持续降低轨迹误差。经过调优的两项损失在简洁损失对比中进一步改善位置重建，但历史复合目标仍具有最低朝向误差。综合而言，这些发现支持将预测表征映射为运动作为实现高效连续无人机导航的一条路径。后续评估应进一步量化未见环境下的任务特定闭环完成情况、物理尺度精度和完整导航更新延迟。除本文研究的提前退出计算之外，推测式推理及其经验尺度规律等模型侧加速机制~\cite{yan2025scaling}也可能成为降低大型预测模型计算成本、同时保持运动质量的一条互补研究方向。
\FloatBarrier

\bibliography{references}
\bibliographystyle{sciencemag}

\end{document}